\pdfoutput=1

\documentclass[11pt]{article}
\usepackage{float}
\usepackage{booktabs}
\usepackage{acl}
\usepackage[utf8]{inputenc}
\usepackage{amsmath}
\usepackage{multirow}
\usepackage{stfloats}
\usepackage{hyperref}
\usepackage{booktabs}
\usepackage{setspace}
\usepackage{array}
\usepackage{mathrsfs}
\usepackage{times}
\usepackage{latexsym}
\usepackage{booktabs}
\usepackage{tabularx}
\usepackage[T1]{fontenc}

\usepackage[utf8]{inputenc}
\usepackage{tabularx}

\usepackage{microtype}

\usepackage{inconsolata}
\usepackage{CJK}

\usepackage{graphicx}
\usepackage{pgfplots}
\usepackage{pgfplotstable}
\pgfplotsset{compat=1.17}
\title{Can Large Language Models Understand DL-Lite Ontologies? An Empirical Study}

 \author{Keyu Wang${}^{1}$, Guilin Qi${}^{1,2\dag}$, Jiaqi Li${}^{1,2}$, Songlin Zhai${}^{1,2}$\\
         ${}^{1}$ School of Computer Science and Engineering, Southeast University, Nanjing, China \\ 
         ${}^{2}$ Key Laboratory of New Generation Artificial Intelligence Technology and Its \\ Interdisciplinary Applications (Southeast University), Ministry of Education, China \\
         \{ky-wang,gqi, jqli, songlin\_zhai\}@seu.edu.cn}

\begin{document}
\maketitle
\begin{abstract}
Large language models (LLMs) have shown significant achievements in solving a wide range of tasks. Recently, LLMs' capability to store, retrieve and infer with symbolic knowledge has drawn a great deal of attention, showing their potential to understand structured information. However, it is not yet known whether LLMs can understand Description Logic (DL) ontologies. In this work, we empirically analyze the LLMs' capability of understanding DL-Lite ontologies covering 6 representative tasks from syntactic and semantic aspects. With extensive experiments, we demonstrate both the effectiveness and limitations of LLMs in understanding DL-Lite ontologies. We find that LLMs can understand formal syntax and model-theoretic semantics of concepts and roles quite well. However, LLMs struggle with understanding TBox NI (Negative Inclusion) transitivity and handling ontologies with large ABoxes. We hope that our experiments and analyses provide more insights into LLMs and inspire to build more faithful knowledge engineering solutions. 
\end{abstract}

\section{Introduction}
Large Language Models (LLMs) \cite{gpt,gpt4,xiaoming} have showcased remarkable proficiency in understanding textual data and revolutionized the field of natural language processing. Recent studies suggest that LLMs possess adaptability to store, retrieve and infer with symbolic knowledge such as knowledge graphs (KGs) \cite{llm4skg, knowledge-solver}, sparking interest in their potential for understanding structured information. However, LLMs' capacity in understanding more complex symbolic knowledge, Description Logic (DL) ontologies, remains unexplored. 

\renewcommand{\thefootnote}{}
\footnotetext{${}^\dag$ Corresponding author. }

\begin{figure}[t]
\begin{center}
\includegraphics[width=0.47\textwidth]{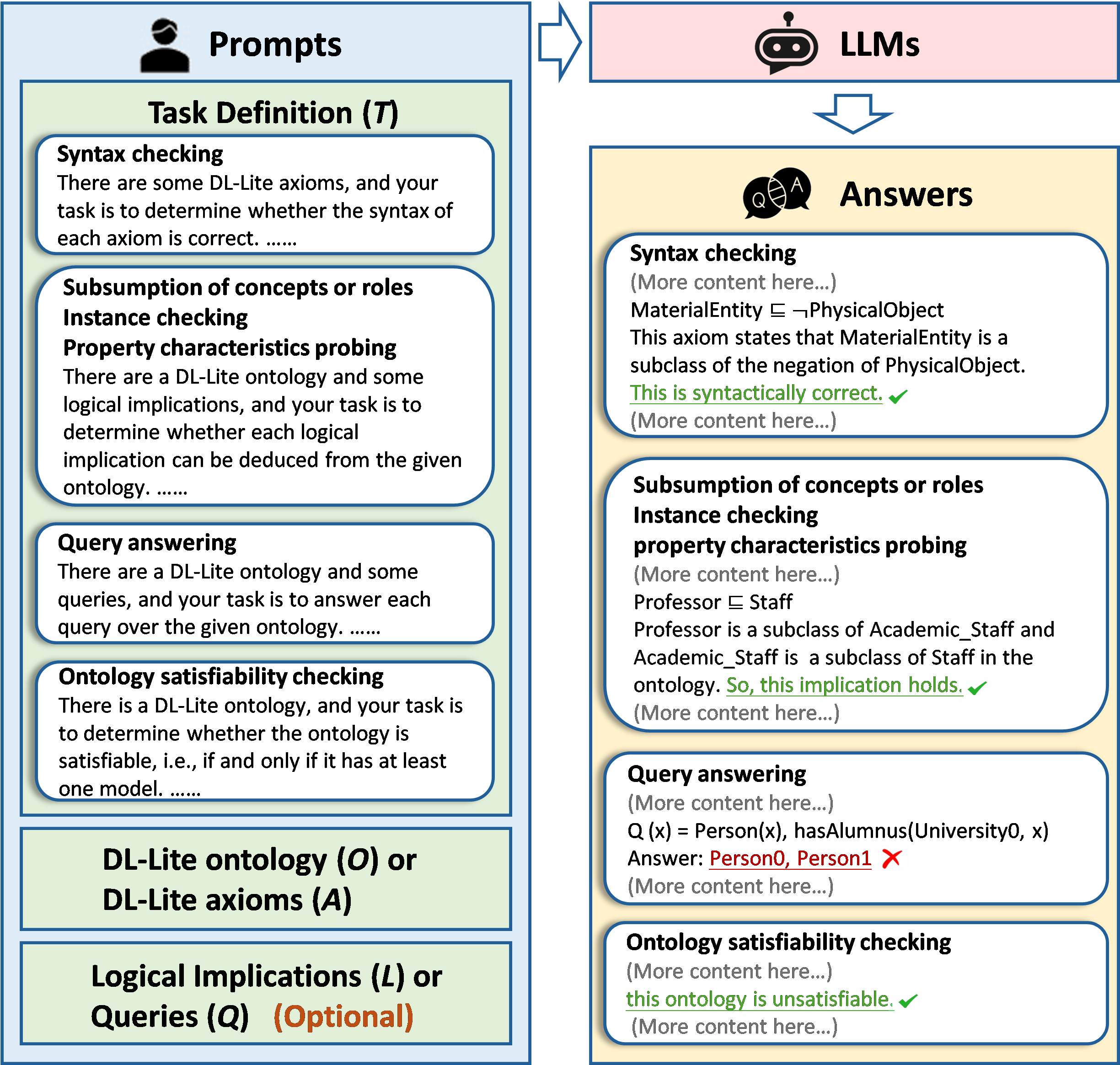}
\end{center}
\caption{Illustration and examples of  evaluation tasks.}
\label{prompt_overview}
\end{figure}

Compared with KGs, DL ontologies have more fined-grained knowledge representation with formal syntax and model-theoretic semantics. For syntax, while most KGs generally only support atomic entities like $PhdStudent$, DL ontologies can support various constructors and compound concepts such as $\neg PhdStudent \sqcap \exists HasStudentID$. For semantics, DL ontologies have model-theoretic semantics. For example, the above complex concept can be interpreted as the set of individuals who are not PhD students but do have a student ID. Further, DL ontologies efficiently support logical reasoning such as $ C_1 \sqsubseteq C_2, C_3 \sqsubseteq \neg  C_2 \rightarrow C_1 \sqsubseteq \neg C_3$. This denotes if $C_1$ is a subclass of $C_2$ and $C_3$ is disjoint with $C_2$, then it logically follows that $C_1$ must also be disjoint from $C_3$ (e.g., $Dog \sqsubseteq Mammal, Bird \sqsubseteq \neg Mammal \rightarrow Dog \sqsubseteq \neg Bird$). Understanding a DL ontology goes beyond just the capabilities of storage, retrieval, and inference, but involves a deeper comprehension of its formal syntax and semantic interpretations.

While the necessity for more detailed investigations for LLMs' capacity in understanding DL ontologies is clear, a comprehensively evaluation presents a challenge. Most related works focus on LLMs' capacity to capturing patterns in KGs \cite{llm4skg, knowledge-solver}, far away from indicating that LLMs possess the ability to understand DL ontologies.  Even though many endeavors study whether LLMs can do logical reasoning \cite{logic1,logic2,logic3,logic4}, few of them explore LLMs' capacity with DL services. DL is primarily focused on representing and reasoning about the hierarchical relationships and properties of concepts within a domain, distinguishing it from other logics by its emphasis on structured, formal ontology. This research gap highlights the significance and challenges in comprehensively evaluating whether LLMs can understand DL ontologies.


In this study, we investigate how effectively LLMs can understand DL-Lite ontologies, a member of the DL ontology family known for simplicity and efficient reasoning.  We present an evaluation framework that comprehensively assesses LLMs' capability to understand DL-Lite ontologies in two aspects, respectively, whether LLMs can grasp the formal representations (the syntactic aspect) and whether LLMs can understand the semantic interpretations of ontologies and  effectively utilize them  (the semantic aspect). For the syntactic aspect, we investigate whether LLMs can comprehend the structural rules, valid statements, and expressions of DL-Lite through syntax checking. For the semantic aspect, we first investigate whether LLMs can understand the semantics of concepts and roles from two aspects, intension and extension, by subsumption of concepts or roles and instance checking respectively. Additionally, we probe property characteristics in DL-Lite ontologies, such as inverse roles and functional roles. Further, we conduct query answering and ontology satisfiability checking to evaluate whether LLMs can understand the semantics of the whole ontologies.  Figure \ref{prompt_overview} gives an illustration  of these tasks. 

Through extensive experiments, we find that: \\
\hspace*{0.2cm} $\bullet$ LLMs possess capacity for understanding DL-Lite syntax (Section \ref{B}). \\
\hspace*{0.2cm} $\bullet$ LLMs can understand the semantics of concepts, roles (Section \ref{A}) and some property characteristics (Section \ref{C}). \\
\hspace*{0.2cm} $\bullet$  LLMs fail to understand some TBox NI transitivity rules, thus LLMs' capability for subsumption of concepts or roles is limited (Section \ref{A}). \\
\hspace*{0.2cm} $\bullet$ LLMs fail to handle ontologies with large scale ABoxes, thus LLMs' capability for instance checking and query answering is limited (Section \ref{A}, Section \ref{D}). \\ 
\hspace*{0.18cm} $\bullet$ LLMs can perform  ontology satisfiability checking with DL-Lite ontologies but struggle with detecting inconsistency in complex ontologies (Section \ref{E}).

To the best of our knowledge, this is the first study to conduct comprehensive evaluations about whether LLMs can understand DL-Lite ontologies. Overall, our work contributes to a better understanding of LLMs’ behaviors and inspires to build more faithful knowledge engineering solutions.

\section{Related Work}
\subsection{LLMs for Syntax Understanding}
With the arrival of LLMs, some works focus on using LLMs to translate natural language into formal language to reduce labor in real-world applications. For example, \citet{LLM4EL} use  ChatGPT  to generate entity relation (ER) diagrams  for conceptual modeling and \citet{LLM4FOL} present a fine-tuned LLaMA-7B model to translate natural language into first-order logic (FOL). \citet{LLM4OWL} convert natural language sentences into OWL Functional Syntax, showing LLMs' prospect of ontology engineering. 
However, there is a significant difference in syntax between DL and other formal languages like ER, FOL and OWL, and few works study whether LLMs can understand DL syntax.

\subsection{LLMs for Semantics Understanding}
Some studies, like \cite{llm4skg, knowledge-solver}, focus on LLMs' capacity of matching up to knowledge that presents in KGs, but such kind of factual knowledge is not the main focus of DL ontology.  \citet{concept-aware} analyze how well LLMs capture concepts and their structures, showing evidence that LLMs can understand conceptual knowledge, but DL ontologies support more automated reasoning than just conceptual taxonomies. Further, recent works conduct evaluations on how effectively LLMs can capture logic and perform logical reasoning \cite{logic1, logic2, logic3, logic4, logic5}. However, none of them study LLMs' capacity in understanding DL semantics. Focusing on representation and reasoning with structured, formal ontology, DL provides  formal semantics based on model theory and strikes a balance between expressiveness and computational tractability , making differences with other logics.

Additionally, some works study LLMs acting as knowledge bases \cite{lmkb}, which focus on LLMs' capacity for storing and retrieving knowledge. In contrast, we conduct an in-depth study of LLMs' understanding of the components (e.g., concepts and roles) in DL ontologies, like how these components get their meanings (from two aspects, extension and intension) and how the meaning of a complex expression depends on its parts (considering various reasoning services). 


\section{Preliminaries} \label{P}
In this section, we briefly recall some basic notions about DL-Lite ontology \cite{dl-lite, dl-lite-onto}. Particularly, we focus on DL-Lite$_{\textit{core}}$, DL-Lite$_\mathcal{F}$ and DL-Lite$_\mathcal{R}$, three members in DL-Lite family, while our evaluation framework can be applied to any other description logics (DLs) such as DL-Lite$_\mathcal{A}$, $\mathcal{ALC}$ and $\mathcal{EL}$.  

\textbf{DL-Lite ontology}. We start from DL-Lite$_{\textit{core}}$ concepts and roles, which are defined as follows: \\
\hspace*{0.4cm} $ B ::= A \ | \ \exists R \ | \ \exists R^{-}  \ \ \ \qquad  R ::= P \ | \ P^{-} $\\
\hspace*{0.4cm}  $ C ::= B \ | \ \neg B \ | \ C_1 \sqcap C_2 \quad E ::= R \ | \ \neg R $ \\
where $A$ denotes an atomic concept, $P$ denotes an atomic role, and $P^-$ denotes the inverse of the atomic role $P$ and $\neg R$ denote the negation of $R$. We call $B,R,C,E$ a basic concept, a basic role, a general concept and a general role respectively.

A DL-Lite$_{\textit{core}}$ ontology $\mathcal{O}=\langle\mathcal{T}, \mathcal{A}\rangle$ consists of a TBox $\mathcal{T}$ and an ABox $\mathcal{A}$.  $\mathcal{T}$ is formed by a finite set of concept inclusion assertions of the form $B \sqsubseteq \mathrm{C}$. $\mathcal{A}$ is formed by a finite set of membership assertions on atomic concepts and on atomic roles, of the form $A(a)$ and $P(a,b)$, where $a$ and $b$ are constants. DL-Lite$_{\mathcal{R}}$ extends DL-Lite$_{\text {core}}$ with role inclusion assertions of the form $R  \sqsubseteq E$ and DL-Lite$_{\mathcal{F}}$ extends DL-Lite$_{\textit{core}}$ with functionality on roles or on their inverses of the form (funct $R$).

The semantics of DL-Lite is given in a model-theoretic way via interpretations over a fixed infinite domain $\Delta$. Given an interpretation $\mathcal{I}$ and an assertion $\alpha$, $\mathcal{I} \models \alpha$ means that $\mathcal{I}$ is a model of $\alpha$. An interpretation is a model of a DL-Lite ontology $\mathcal{O}$, if and only if it is a model for each assertion in $\mathcal{O}$. An ontology $\mathcal{O}$ is satisfiable if it has at least one model. $\mathcal{O}$ logically implies an assertion $\alpha$, written $\mathcal{O} \models \alpha$, if all models of $\mathcal{O}$ are also models of $\alpha$.

\textbf{Reasoning services with DL-Lite ontology.} Designed for knowledge representation and efficient reasoning, DL-Lite ontology supports several DL reasoning services \cite{dl-lite}: \\
- Ontology satisfiability checking: given an ontology $\mathcal{O}$, verify whether $\mathcal{O}$ admits at least one model; \\
- Logical implication of $\mathcal{O}$ assertions, which consists of the following sub-problems: \\
\hspace*{0.2cm} $\bullet$ Instance checking: given an ontology $\mathcal{O}$, a concept $C$ and a constant $a$ (resp., a role $R$ and a pair of constants $a$ and $b$ ), verify whether $\mathcal{O} \models C(a)$ (resp., $\mathcal{O} \models R(a, b))$. \\
\hspace*{0.2cm} $\bullet$ Subsumption of concepts or roles: given a TBox $\mathcal{T}$ and two general concepts $C_1$ and $C_2$ (resp., two general roles $R_1$ and $R_2$ ), verify whether $\mathcal{T} \models C_1 \sqsubseteq C_2$ (resp., $\mathcal{T} \models R_1 \sqsubseteq R_2$ ). \\
\hspace*{0.2cm} $\bullet$ Checking functionality - given a TBox $\mathcal{T}$ and a basic role $R$, verify whether $\mathcal{T} \models($ funct $R)$. \\
- Query answering: given an ontology $\mathcal{O}$ and a query $q$  over $\mathcal{O}$, compute the answer set $\operatorname{ans}(q, \mathcal{O})$.

A key characteristic of DL-Lite syntax and semantics is that they are primarily designed for performing these DL reasoning services efficiently. Conducting an extensive evaluation of LLMs for these tasks is beneficial to provide insights into whether LLMs can understand DL-Lite ontologies.   

\textbf{Transitivity rules.} For instance checking and subsumption of concepts or roles, we especially focus on deducing logical implications with some reasoning rules. Borrowing the idea of Canonical Interpretation (PI-closure) and Closure of Negative Inclusion Assertions (NI-closure) from \cite{dl-lite, dl-lite-onto}, we collect the reasoning rules in three categories, 2 TBox PI (positive inclusion) transitivity rules, 11 TBox NI (negative inclusion) transitivity rules and 5 ABox transitivity rules. We cover them in Appendix \ref{AA} and there are some examples  below: \\
\noindent \fbox{\small{\parbox{\linewidth}{%
TBox PI transitivity examples:\\
$\alpha=C_1 \sqsubseteq C_2,   \beta=C_2 \sqsubseteq C_3   \rightarrow  \beta_{\text{new}}=C_1 \sqsubseteq C_3$\\
$\alpha=R_1 \sqsubseteq R_2 ,  \beta=R_2 \sqsubseteq R_3   \rightarrow  \beta_{\text{new}}=R_1 \sqsubseteq R_3$\\ 
TBox NI transitivity examples:\\
$\alpha=C_1 \sqsubseteq C_2,  \beta=C_3 \sqsubseteq \neg C_2 \rightarrow \beta_{\text {new }}=C_1 \sqsubseteq \neg C_3 \\
 \alpha=R_1 \sqsubseteq R_2, \beta=\exists R_2^{-} \sqsubseteq \neg C \rightarrow \beta_{\text {new }}=\exists R_1^{-} \sqsubseteq \neg C $\\ 
ABox transitivity examples: \\ 
$  \alpha=C \sqsubseteq \exists R , \beta=C(a)  \rightarrow  \beta_{\text {new }}=R\left(a, a_{\text {new }}\right) \\
  \alpha=\exists R \sqsubseteq C, \beta=R\left(a, a^{\prime}\right)  \rightarrow  \beta_{\text {new}}=C(a) $
}}}


\begin{figure*}[htbp]
\begin{center}
\includegraphics[width=1.00\textwidth]{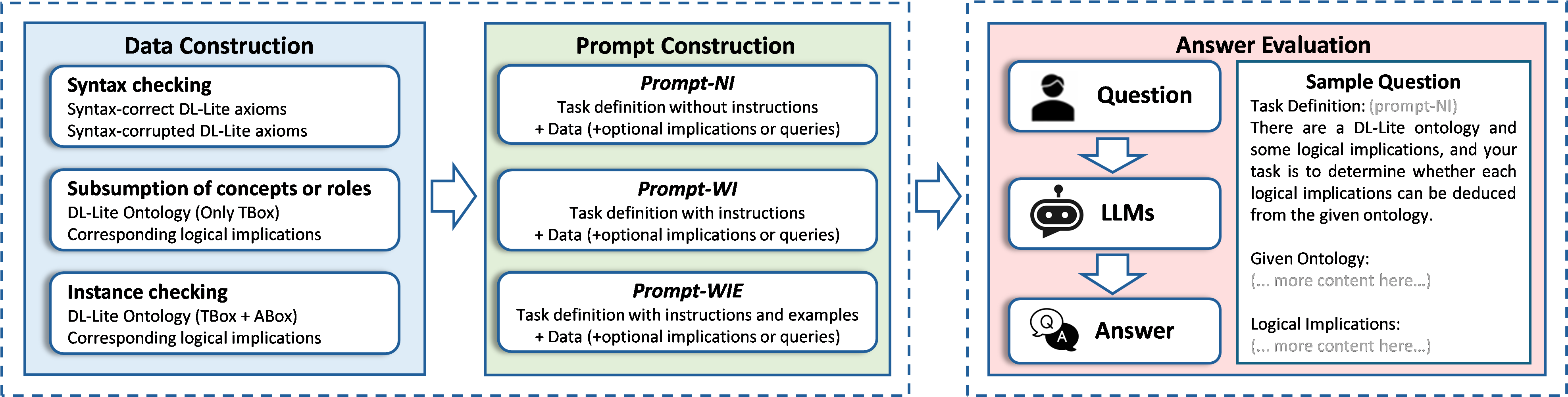}
\end{center}
\caption{Evaluation pipeline for syntax checking, subsumption of concepts or roles, and instance checking.}
\label{overview}
\end{figure*}

\begin{table*}
\centering
\scalebox{0.55}{
\begin{tabular}{c|ccc|ccc|ccc|ccc|ccc}
\hline Datasets & \multicolumn{3}{c|}{GO} & \multicolumn{3}{c|}{FMA} & \multicolumn{3}{c|}{MarineTLO} & \multicolumn{3}{c|}{Music} & \multicolumn{3}{c}{OBI} \\ 
\hline Metric  & Precision & Recall & F1-Score  & Precision & Recall & F1-Score & Precision & Recall & F1-Score  & Precision & Recall & F1-Score  & Precision & Recall & F1-Score \\ \hline
GPT3.5--\textit{NI} & 66 & 90 & 76 & 100 & 100 & 100 & 96 & 87 & 91 & 100 & 97 & 98 & 100 & 100 & 100 \\ 
GPT3.5--\textit{WI} & 66 & 97 & 79   & 68  & 100  & 81  & 100 & 100 & 100 & 83 & 100  & 91 & 83  & 97 & 89 \\ 
GPT3.5--\textit{WIE} & 72 & 93 & 81 & 65 & 100 & 79 & 87 & 87 & 87 & 63 & 100 & 77 & 82 & 90 & 86 \\ 
GPT4o--\textit{NI} & 100 & 97 & 98 & 86 & 100 & 92 & 100 & 100 & 100 & 100 & 100 & 100 & 100 & 97 & 98\\ 
GPT4o--\textit{WI}  & 91 & 100  & 95  & 88  & 100 &  94 & 97 & 100 & 98 & 100 & 100 &  100 & 79 &  100  & 88 \\ 
GPT4o--\textit{WIE} & 100 & 100 & 100 & 88 & 100 & 94 & 97 & 100 & 98 & 97 & 100 & 98 & 94 & 100 & 97\\ 
LLaMA3-8b--\textit{NI} & 65 & 93 & 77 & 50 & 100 & 67 & 50 & 100 & 67 & 63 & 97 & 76 & 58 & 93 & 71 \\ 
LLaMA3-8b--\textit{WI} & 91 & 100 & 95 & 50 & 100 & 67 & 67 & 100 & 80 & 76 & 97 & 85 & 64 & 90 & 75 \\ 
LLaMA3-8b--\textit{WIE} & 67 & 93 & 78 & 58 & 97 & 73 & 63 & 100 & 77 & 78  & 97 & 86 & 71 & 100 & 83 \\ 
\hline
\end{tabular}}
\caption{Performances of LLMs in syntax checking (\%).}
 \label{tab4}
\end{table*}

\section{Unveiling LLMs' Capabilities in Understanding DL-Lite Ontology}
In this section, we comprehensively investigate how effectively LLMs can understand DL-Lite ontologies, especially, grasp the formal representations (syntax) and interpretations of elements in ontologies (semantics). We conduct a series of tasks, including syntax checking, subsumption of concepts or roles, instance checking, query answering, ontology satisfiability checking and property characteristics probing. Figure \ref{overview} presents an overview of the evaluation framework for the first three tasks. We collect specified datasets for each task and construct three prompts of binary questions, and test three LLMs, namely, GPT3.5 \cite{gpt3.5-1}, GPT4o\footnote{https://openai.com/index/hello-gpt-4o/} \cite{gpt4} and LLaMA3-8B\footnote{https://github.com/meta-llama/llama3} \cite{llama}.  The evaluation pipelines of the other three tasks introduced later are quite similar. All the data for evaluation is released at \href{https://github.com/keyu-wang-2002/Can-LLMs-Understand-DL-Lite-Ontologies}{https://github.com/keyu-wang-2002/Can-LLMs-Understand-DL-Lite-Ontologies}.

\subsection{Can LLMs Understand the Syntax of DL-Lite Ontologies?} \label{B}
An important aspect of how effectively LLMs can understand DL-Lite ontologies is their capacity to comprehend  the syntax. In this section, we conduct syntax checking to evaluate LLMs’ comprehension of structural rules and the construction of valid statements and expressions in DL-Lite ontologies. 

\textbf{Datasets.} We select several commonly used DL ontologies, including Gene Ontology (GO) \cite{go}, Foundational Model of Anatomy (FMA) \cite{fma}, Ontology for Biomedical Investigations (OBI) \cite{obi}, MarineTLO \cite{marine} and the Music Ontology \cite{music}. For each DL ontology, we randomly collect 30 DL-Lite axioms. For each collected axiom, we insert random one type of syntax error, such as invalid quantifier (eg. $\exists TeachesTo \rightarrow \exists\exists TeachesTo$) and invalid conjunction (eg. $ Professor \sqcap \exists TeachesTo \rightarrow \sqcap \ Professor \ \exists TeachesTo$). We summarize typical syntax errors in DL-Lite in Appendix \ref{AB}. We build 150 correct and 150 corrupted DL-Lite axioms as datasets for syntax checking.

\textbf{Experimental setup.} We utilize binary questions for syntax checking.  Generally, the prompts include task description (\textit{T}) and the input DL-Lite axioms (\textit{A}). We design three kinds of prompts: \\
$\bullet$ prompt without any instructions about DL-Lite syntax in \textit{T}, denoted as \textit{NI} (\underline{N}o \underline{I}nstructions);\\ 
$\bullet$ prompt with instructions about DL-Lite syntax in \textit{T}, denoted as \textit{WI} (\underline{W}ith \underline{I}nstructions);\\ 
$\bullet$ prompt with instructions about DL-Lite syntax and corresponding examples in \textit{T}, denoted as \textit{WIE} (\underline{W}ith \underline{I}nstructions and \underline{E}xamples). 

Figure \ref{prompt_overview} shows an example and we cover detailed prompts in Appendix \ref{AC}.

\textbf{Results analysis.} In Table \ref{tab4}, we present precision, recall and F1 score of tested LLMs and prompts. Overall, LLMs possess the ability to understand DL-Lite syntax. We find that no matter what kinds of prompts we use, GPT4o achieves good results on all the five data sources. In comparison, LLaMA3-8b shows relatively poor results. To deliver a more in-depth investigation, we conduct analyses for the following questions:

\textbf{\textit{Can instructions or examples benefit LLMs' understanding of DL-Lite syntax?}} For GPT3.5 and GPT4o, there is little difference among the three prompts, while performances of LLaMA3-8B--\textit{WI} and LLaMA3-8B--\textit{WIE} are significantly better than those of LLaMA3-8B--\textit{NI}. This may be because GPT3.5 and GPT4o have learned detailed DL-Lite syntax during training but LLaMA3-8B hasn't. 

\textbf{\textit{What types of errors do LLMs usually make for syntax checking?}} In most cases, LLMs achieve high recall and relatively low precision, since LLMs hardly mistake correct axioms, but do sometimes treat incorrect axioms as correct. Especially, we find that LLMs sometimes perform poorly in distinguishing between concepts and roles. For example, they may treat $\exists isConnectedTo \sqsubseteq Organ^{-}$ as syntax-correct, which is incorrect since inverse ($^{-}$) can only be put on roles.

\subsection{Can LLMs Understand the Semantics of DL-Lite ontologies?}
Another aspect of whether LLMs can understand ontologies is their capacity to comprehend the semantics. Semantics goes beyond the syntactic structure and explores the interpretation and significance of the elements like concepts and roles of the ontology. In this section, we explore the capability of LLMs to understand the semantics of the components within ontology (i.e., concepts and roles)   considering instance checking and subsumption of concepts or roles. Additionally, we probe some property characteristics (i.e., inverse roles and functional roles) in DL-Lite ontologies. Further, we conduct query answering and ontology satisfiability checking to explore LLMs' capacity to understand the semantics of the whole ontologies.

\subsubsection{Semantics of Concepts and Roles} \label{A}
We evaluate the capacity of LLMs to understand the semantics of concepts and roles from two aspects: extension and intension \cite{exin1, exin2, exin3, eike}. The extension of a concept or role refers to the set of individuals or objects that fall under that concept or role \cite{exin1, exin3}. For example, the extension of the concept ``President of the U.S.'' would be the set of all individuals considered as U.S. presidents such as ``Barack Obama'' and ``Joe Biden''.  The intension of a concept or role refers to the characteristics, properties, or conditions that determine whether an individual belongs to  that concept or role \cite{exin3}. For example, ``President of the U.S.'' is a ``Politician'' and ``someone who plays a role in federal legislation''\footnote{\href{https://en.wikipedia.org/wiki/President\_of\_the\_United\_States}{en.wikipedia.org/wiki/President\_of\_the\_United\_States}}.

We use instance checking for the former, since it involves determining whether a particular individual  belongs to a specified concept within a given ontology. Subsumption of concepts or roles is for the latter, which involves determining whether one concept or role is subsumed by another more general concept or role, reflecting the attributes, characteristics, constraints, and conditions encompassed by the inherent intension. 

\begin{table}[H]

\centering
\scalebox{0.55}{
\begin{tabular}{ccccc}
\toprule
Data Sources& \#T. $B\sqsubseteq C$ & \#T. $R \sqsubseteq E$& \#L. $B \sqsubseteq C$ & \#L. $R \sqsubseteq E$\\
\midrule
VICODI & 193 & 9 & 195 & 9\\
STOCKEXCHANGE & 26 & 0 & 12 & 0\\
UNIVERSITY & 36 & 5 & 31 & 9\\
ADOLENA & 100 & 0 & 72 & 0\\
SEMINTEC & 55 & 0 & 47 & 0\\
\bottomrule
\end{tabular}}
\caption{Statistics about data sources for subsumption of concepts or roles. \# denotes ``the number of'', and T., L. denote TBox and logical implications respectively. }
\label{tab1}
\end{table}

\begin{table}[H]

\centering
\scalebox{0.65}{
\begin{tabular}{ccccc}
\toprule
Data Sources& \#O. $C(a)$ &\#O. $R(a, b)$ &\#L. $C(a)$ &\#L. $R(a, b)$\\
\midrule
UOBM1 & 2338& 0 &478 & 0 \\
UOBM2 & 1389& 0 &278 & 0 \\
UOBM3 & 678& 0 &136 & 0 \\
UOBM4 & 576& 0 &113 & 0 \\
UOBM5 & 466& 0 &93 & 0 \\
\bottomrule
\end{tabular}}
\caption{Statistics about data sources for instance checking.  \# denotes ``the number of'', and O., L. denote ontology and logical implications respectively.}
\label{tab2}
\end{table}

\definecolor{color1}{RGB}{101,39,39}
\definecolor{color2}{RGB}{220,175,175}
\definecolor{color3}{RGB}{243,175,250}
\definecolor{color4}{RGB}{180,175,250}
\definecolor{color5}{RGB}{158,174,176}
\definecolor{color6}{RGB}{83,146,121}
\definecolor{color7}{RGB}{75,118,75}
\definecolor{color8}{RGB}{212,215,135}
\definecolor{color9}{RGB}{215,169,135}
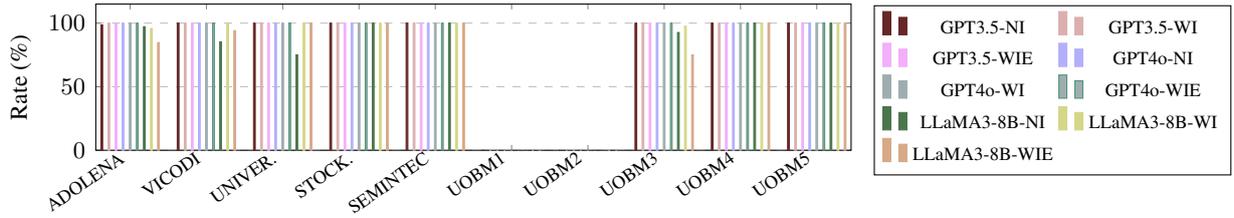
\begin{figure*}
\centering
\begin{tikzpicture}
    \begin{axis}[
        ybar,
        width=0.72\textwidth,
        height=0.22\textwidth,
        bar width=0.65pt,
        ylabel={Rate (\%)}, 
        ylabel style={yshift=-0.1em,font=\footnotesize},
        symbolic x coords={ADOLENA, VICODI, UNIVER., STOCK., SEMINTEC, UOBM1, UOBM2, UOBM3, UOBM4, UOBM5},
        xtick=data,
        x tick label style={rotate=35, anchor=east, font=\scriptsize},
        y tick label style={font=\small},
        tick align = inside,
        ymin=0, ymax=100,
        ymajorgrids=true,  
        grid style=dashed,
        tick label style={font=\Large},
        label style={font=\small},
        tick style={major tick length=2pt, minor tick length=1pt},
        legend style={
            at={(1.03,0.41)},  
            anchor=west,
            font=\scriptsize,
            legend columns=2  
        },
        enlarge x limits=0.05,
        enlarge y limits={value=0.15, upper}  
    ]
    
     \addplot[draw=color1, fill=color1] coordinates {(ADOLENA, 98.6) (VICODI, 100.00) (UNIVER., 100.00) (STOCK., 100.00) (SEMINTEC, 100.00) (UOBM1, 0.00) (UOBM2, 0.00) (UOBM3, 100.00) (UOBM4, 100.00) (UOBM5, 100.00)};
    \addplot[draw=color2, fill=color2] coordinates {(ADOLENA, 100.00) (VICODI, 100.00) (UNIVER., 100.00) (STOCK., 100.00) (SEMINTEC, 100.00)(UOBM1, 0.00) (UOBM2, 0.00) (UOBM3, 100.00) (UOBM4, 100.00) (UOBM5, 100.00)};
    \addplot[draw=color3, fill=color3] coordinates {(ADOLENA, 100.00) (VICODI, 100.00) (UNIVER., 100.00) (STOCK., 100.00) (SEMINTEC, 100.00)(UOBM1, 0.00) (UOBM2, 0.00) (UOBM3, 100.00) (UOBM4, 100.00) (UOBM5, 100.00)};
    \addplot[draw=color4, fill=color4] coordinates{(ADOLENA, 100.00) (VICODI, 100.00) (UNIVER., 100.00) (STOCK., 100.00) (SEMINTEC, 100.00)(UOBM1, 0.00) (UOBM2, 0.00) (UOBM3, 100.00) (UOBM4, 100.00) (UOBM5, 100.00)};
    \addplot[draw=color5, fill=color5] coordinates{(ADOLENA, 100.00) (VICODI, 100.00) (UNIVER., 100.00) (STOCK., 100.00) (SEMINTEC, 100.00)(UOBM1, 0.00) (UOBM2, 0.00) (UOBM3, 100.00) (UOBM4, 100.00) (UOBM5, 100.00)};
    \addplot[draw=color6, fill=color5] coordinates {(ADOLENA, 100.00) (VICODI, 100.00) (UNIVER., 100.00) (STOCK., 100.00) (SEMINTEC, 100.00)(UOBM1, 0.00) (UOBM2, 0.00) (UOBM3, 100.00) (UOBM4, 100.00) (UOBM5, 100.00)};
    \addplot[draw=color7, fill=color7] coordinates {(ADOLENA, 97.2) (VICODI, 85.3) (UNIVER., 75.00) (STOCK., 100.00) (SEMINTEC, 100.00)(UOBM1, 0.00) (UOBM2, 0.00) (UOBM3, 92.65) (UOBM4, 100.00) (UOBM5, 100.00)};
    \addplot[draw=color8, fill=color8] coordinates {(ADOLENA, 95.80) (VICODI, 99.50) (UNIVER., 100.00) (STOCK., 100.00) (SEMINTEC, 100.00)(UOBM1, 0.00) (UOBM2, 0.00) (UOBM3, 97.61) (UOBM4, 100.00) (UOBM5, 100.00)};
    \addplot[draw=color9, fill=color9] coordinates {(ADOLENA, 84.70) (VICODI, 94.10) (UNIVER., 100.00) (STOCK., 100.00) (SEMINTEC, 100.00)(UOBM1, 0.00) (UOBM2, 0.00) (UOBM3, 75.00) (UOBM4, 100.00) (UOBM5, 100.00)};
 
    \legend{GPT3.5-NI, GPT3.5-WI, GPT3.5-WIE, GPT4o-NI, GPT4o-WI, GPT4o-WIE, LLaMA3-8B-NI, LLaMA3-8B-WI, LLaMA3-8B-WIE}
    \end{axis}
\end{tikzpicture}
\caption{Performances of LLMs in subsumption of concepts or roles and instance checking.}
\label{sub}
\end{figure*}

\begin{table*}[ht] \scriptsize
	\centering
	\begin{tabularx}{\textwidth}{X|X}
		\toprule
		  \textbf{DL-Lite Ontology} & \textbf{Logical Implications}  \\
		\midrule
  \textbf{Case 1}: TBox $=$ \{ $C_1 \sqsubseteq C_2$, $C_2 \sqsubseteq \neg C_3$, $C_4 \sqsubseteq \neg C_2$, $R_1 \sqsubseteq  R_2$, $\exists R_2 \sqsubseteq  \neg C_5$, $C_6 \sqsubseteq  \neg \exists R_2$, $R_3 \sqsubseteq R_4$, $\exists R_4^{-} \sqsubseteq  \neg C_7$, $C_8 \sqsubseteq \neg \exists R_4^{-}$, $R_5 \sqsubseteq R_6$, $R_6 \sqsubseteq \neg R_7$, $R_8 \sqsubseteq \neg R_6$ \}  
   &  $C_1 \sqsubseteq \neg C_3$, $C_1 \sqsubseteq \neg C_4$, $\exists R_1 \sqsubseteq  \neg C_5$, $\exists R_1 \sqsubseteq  \neg C_6$, $\exists R_3^{-} \sqsubseteq  \neg C_7$, $\exists R_3^{-} \sqsubseteq  \neg C_8$, $R_5 \sqsubseteq \neg R_7$, $R_5 \sqsubseteq \neg R_8$ \\ \hline
   \textbf{Case 2}: TBox $=$ \{ $C_1 \sqsubseteq C_2$, $C_1 \sqsubseteq C_3$, $C_2 \sqsubseteq C_4$, $R_1 \sqsubseteq R_2$, $R_3 \sqsubseteq R_4$, $C_5 \sqsubseteq \exists R_5$,  $ \exists R_6 \sqsubseteq C_6 $, $ \exists R_7 \sqsubseteq \exists R_8 $\};
ABox $=$ \{ $C_1(a)$, $C_1(b)$, $R_1(c, d)$, $R_3(e, f)$, $C_5(a)$, $R_6(a, k)$, $R_7(g, h)$\}
& $C_2(a)$, $C_3(a)$, $C_2(b)$, $C_3(b)$, $R_2(c, d)$, $R_4(e, f)$, $C_4(a)$, $C_4(b)$, $R_5(a, \_)$, $C_6(a)$, $R_8(h, \_)$ \\ \hline
\textbf{Case 3}:  TBox $=$ \{$C_1 \sqsubseteq C_2$,  $C_1 \sqsubseteq C_4$, $C_1 \sqsubseteq C_6$, $C_2 \sqsubseteq C_3$, $C_4 \sqsubseteq \neg C_5$, $C_7 \sqsubseteq \neg C_6$, 
$R_1 \sqsubseteq  R_2$, $R_4 \sqsubseteq  R_5$, 
$R_6 \sqsubseteq  R_7$, $R_2 \sqsubseteq  R_3$, 
$\exists R_2 \sqsubseteq \neg C_8$, $ C_9 \sqsubseteq \neg \exists R_2$, $ C_{10} \sqsubseteq \neg \exists R_5^{-}$, $ \exists R_5^{-} \sqsubseteq \neg  C_{11}$, $R_7 \sqsubseteq \neg R_8$, $R_9 \sqsubseteq \neg R_7$, $R_{10} \sqsubseteq \neg R_{10}$, $\exists R_{11} \sqsubseteq \neg \exists R_{11}$, $\exists R_{12}^{-} \sqsubseteq \neg \exists R_{12}^{-}$\}
		& $ C_1 \sqsubseteq C_3 $,  $C_1 \sqsubseteq \neg C_5$, $C_1 \sqsubseteq \neg C_7$, 
  $R_1 \sqsubseteq R_3$, $\exists R_1 \sqsubseteq \neg C_8$, $\exists R_1 \sqsubseteq \neg C_9$, $\exists R_4^{-} \sqsubseteq \neg C_{10}$, $\exists R_4^{-} \sqsubseteq \neg C_{11}$, 
  $ R_6 \sqsubseteq \neg R_8$,
  $ R_6 \sqsubseteq \neg R_9$,
  $\exists R_{10} \sqsubseteq \neg \exists R_{10}$,
  $\exists R_{10}^{-} \sqsubseteq \neg \exists R_{10}^{-}$,
  $R_{11} \sqsubseteq \neg R_{11}$,
  $\exists R_{11}^{-} \sqsubseteq \neg \exists R_{11}^{-}$,
  $ R_{12} \sqsubseteq \neg R_{12}$,
  $\exists R_{12} \sqsubseteq \neg \exists R_{12}$. 
  \\ \hline
  \textbf{Case 4}:  TBox $=$ \{ $C_1 \sqsubseteq C_2$, $C_1 \sqsubseteq \exists R_1$, $\exists R_2 \sqsubseteq C_3$, $\exists R_3 \sqsubseteq \exists R_4$, $R_5 \sqsubseteq R_5$\}; ABox $=$ \{$C_1(a)$, $C_1(b)$, $R_2(c, d)$, $R_3(e, f)$, $R_5(g, h)$\}	&  $C_2(a)$, $R_1(b, \_)$, $C_3(c)$, $R_4(e, \_)$, $R_6(g, h)$.
  \\ \hline
 \textbf{Case 5}:  TBox $=$ \{$C_1 \sqsubseteq C_2$, $C_2 \sqsubseteq C_3$, $C_3 \sqsubseteq C_4$, $C_4 \sqsubseteq C_5$, $C_3 \sqsubseteq C_6$, $C_6 \sqsubseteq C_7$, $R_1 \sqsubseteq R_2$, $R_2 \sqsubseteq R_3$, $R_3 \sqsubseteq R_4$\}     &   $C_1 \sqsubseteq C_3$, $C_1 \sqsubseteq C_4$, $C_1 \sqsubseteq C_5$,  $C_1 \sqsubseteq C_6$, $C_1 \sqsubseteq C_7$,  $C_2 \sqsubseteq C_3$, $C_2 \sqsubseteq C_4$, $C_2 \sqsubseteq C_5$,  $C_2 \sqsubseteq C_6$, $C_2 \sqsubseteq C_7$,  $C_3 \sqsubseteq C_5$,   $C_3 \sqsubseteq C_6$,  $C_3 \sqsubseteq C_7$, $R_1 \sqsubseteq R_3$, $R_1 \sqsubseteq R_4$, $R_2 \sqsubseteq R_4$.
 \\ 
		\bottomrule
	\end{tabularx}
	\caption{Handcrafted ontologies in case study of  transitivity rules.}
    \label{transitivity}
\end{table*} 

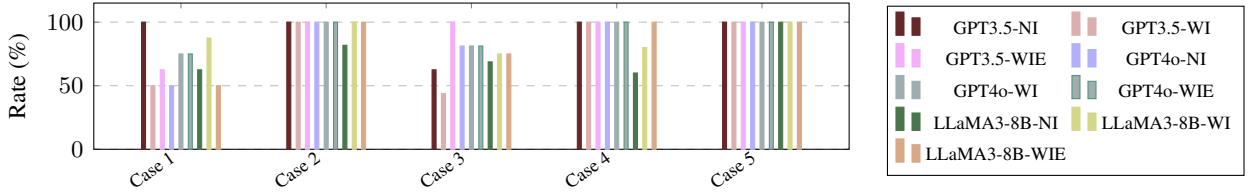
\begin{figure*}[h]
\centering
\begin{tikzpicture}
    \begin{axis}[
        ybar,
        width=0.69\textwidth,
        height=0.22\textwidth,
        bar width=1.50pt,
        ylabel={Rate (\%)}, 
        ylabel style={yshift=-0.1em,font=\footnotesize},
        symbolic x coords={Case 1, Case 2, Case 3, Case 4, Case 5},
        xtick=data,
        x tick label style={rotate=35, anchor=east, font=\scriptsize},
        y tick label style={font=\small},
        tick align = inside,
        ymin=0, ymax=100,
        ymajorgrids=true,  
        grid style=dashed,
        tick label style={font=\Large},
        label style={font=\small},
        tick style={major tick length=2pt, minor tick length=1pt},
        legend style={
            at={(1.05,0.4)},  
            anchor=west,
            font=\scriptsize,
            legend columns=2  
        },
        enlarge x limits=0.15,
        enlarge y limits={value=0.15, upper}  
    ]
    
     \addplot[draw=color1, fill=color1] coordinates {(Case 1, 100.00) (Case 2, 100.00) (Case 3, 62.50) (Case 4, 100.00) (Case 5, 100.00)};
    \addplot[draw=color2, fill=color2] coordinates {(Case 1, 50.00) (Case 2, 100.00) (Case 3, 43.75) (Case 4, 100.00) (Case 5, 100.00)};
    \addplot[draw=color3, fill=color3] coordinates {(Case 1, 62.50) (Case 2, 100.00) (Case 3, 100.00) (Case 4, 100.00) (Case 5, 100.00)};
    \addplot[draw=color4, fill=color4] coordinates{(Case 1, 50.00) (Case 2, 100.00) (Case 3, 81.25) (Case 4, 100.00) (Case 5, 100.00)};
    \addplot[draw=color5, fill=color5] coordinates{(Case 1, 75.00) (Case 2, 100.00) (Case 3, 81.25) (Case 4, 100.00) (Case 5, 100.00)};
    \addplot[draw=color6, fill=color5] coordinates {(Case 1, 75.00) (Case 2, 100.00) (Case 3, 81.25) (Case 4, 100.00) (Case 5, 100.00)};
    \addplot[draw=color7, fill=color7] coordinates {(Case 1, 62.50) (Case 2, 81.81) (Case 3, 68.75) (Case 4, 60.00) (Case 5, 100.00)};
    \addplot[draw=color8, fill=color8] coordinates {(Case 1, 87.50) (Case 2, 100.00) (Case 3, 75.00) (Case 4, 80.00) (Case 5, 100.00)};
    \addplot[draw=color9, fill=color9] coordinates {(Case 1, 50.00) (Case 2, 100.00) (Case 3, 75.00) (Case 4, 100.00) (Case 5, 100.00)};

    \legend{GPT3.5-NI, GPT3.5-WI, GPT3.5-WIE, GPT4o-NI, GPT4o-WI, GPT4o-WIE,LLaMA3-8B-NI, LLaMA3-8B-WI, LLaMA3-8B-WIE}
    \end{axis}
\end{tikzpicture}
\caption{Performances for case study of tranitivity rules.}
\label{transitivity-results}
\end{figure*}

\begin{figure*}[h]
\begin{tikzpicture}
    \begin{axis}[
        ybar,
        width=0.69\textwidth,
        height=0.22\textwidth,
        bar width=3.00pt,
        ylabel={Rate (\%)}, 
        ylabel style={yshift=-0.1em,font=\footnotesize},
        symbolic x coords={Aca.-inv., ecom.-inv., lib.-inv., soc.-inv., med.-inv., Aca.-fun., ecom.-fun., lib.-fun., soc.-fun., med.-fun.},
        xtick=data,
        x tick label style={rotate=35, anchor=east, font=\scriptsize},
        y tick label style={font=\small},
        tick align = inside,
        ymin=0, ymax=100,
        ymajorgrids=true,  
        grid style=dashed,
        tick label style={font=\Large},
        label style={font=\small},
        tick style={major tick length=2pt, minor tick length=1pt},
        legend style={
            at={(1.05,0.4)},  
            anchor=west,
            font=\scriptsize,
            legend columns=1  
        },
        enlarge x limits=0.15,
        enlarge y limits={value=0.15, upper}  
    ]

     \addplot[draw=color3, fill=color3] coordinates {(Aca.-inv., 100.00) (ecom.-inv., 100.00) (lib.-inv., 100.00) (soc.-inv., 100.00) (med.-inv., 100.00) (Aca.-fun., 100.00) (ecom.-fun., 100.00) (lib.-fun., 100.00) (soc.-fun., 100.00) (med.-fun., 100.00)};
     \addplot[draw=color4, fill=color4] coordinates {(Aca.-inv., 100.00) (ecom.-inv., 100.00) (lib.-inv., 100.00) (soc.-inv., 100.00) (med.-inv., 100.00) (Aca.-fun., 100.00) (ecom.-fun., 100.00) (lib.-fun., 100.00) (soc.-fun., 100.00) (med.-fun., 100.00)};
     \addplot[draw=color9, fill=color9] coordinates {(Aca.-inv., 58.33) (ecom.-inv., 100.00) (lib.-inv., 42.86) (soc.-inv., 100.00) (med.-inv., 84.62) (Aca.-fun., 62.50) (ecom.-fun., 100.00) (lib.-fun., 100.00) (soc.-fun., 100.00) (med.-fun., 100.00)};
 
    \legend{GPT3.5-NI, GPT4o-NI, LLaMA3-8B-NI}
    \end{axis}
\end{tikzpicture}
\caption{Performances for probing of inverse role property and (inverse) functional role property.}
\label{property}
\end{figure*}
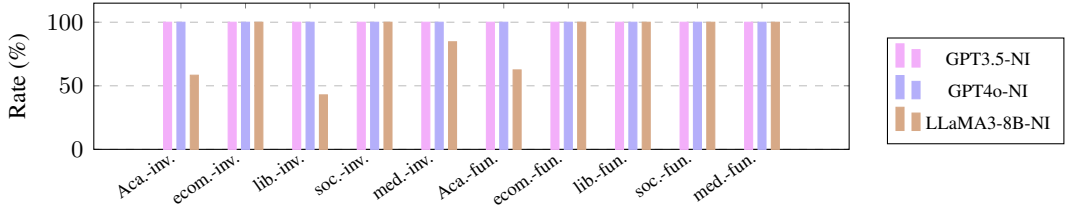

\textbf{Datasets.} For subsumption of concepts or roles, we use the TBox of existing DL-Lite ontologies. We select 4 DL-Lite$_\mathcal{R}$ ontologies, VICODI \cite{vicodi}, STOCKEXCHANGE \cite{stockexechange}, UNIVERSITY \cite{lubm}, ADOLENA \cite{adolena} from \cite{data1}, and SEMINTEC from \cite{data2} as approximation of DL-Lite ontology. For instance checking, we construct a series of DL-Lite ontologies of varying sizes using the UOBM benchmark \cite{uobm}.  We select a variant of UOBM ontology, denoted as UOBM0, and derive five additional ontologies with significantly different ABox sizes by randomly removing class assertions from UOBM0, which are labeled as UOBM1, UOBM2, UOBM3, UOBM4 and UOBM5 respectively.  

Then we load the ontologies into Protégé \footnote{https://protege.stanford.edu/} and utilize the reasoning engine HermiT \cite{hermit} to infer logical implications.  
We cover the details of using Protégé to obtain logical implications in Appendix \ref{AD}. Because there are a large number of logical implications in instance checking, we randomly select a subset for evaluation. Table \ref{tab1} and Table \ref{tab2} show the statistical details.

\textbf{Experimental setup.} The prompts include task description (\textit{T}), input ontology (\textit{O}, only TBox for subsumption of concepts or roles while TBox $+$ ABox for instance checking) and logical implications (\textit{L}).  We design three kinds of prompts: \\
$\bullet$ prompt without any instructions about reasoning rules  in \textit{T}, denoted as \textit{NI};\\ 
$\bullet$ prompt with instructions about reasoning rules (TBox PI transitivity, TBox NI transitivity for concept or role subsumption, and ABox transitivity for instance checking) in \textit{T}, denoted as \textit{WI};\\ 
$\bullet$ prompt with instructions about reasoning rules (same as above) and corresponding examples in \textit{T}, denoted as \textit{WIE}. 

Figure \ref{prompt_overview} shows examples and we cover detailed prompts in Appendix \ref{AC}. The evaluation metric is the ratio of logical implications that LLMs can deduce to all the logical implications.

\textbf{Results analysis.} The performances of LLMs in subsumption of concepts or roles and instance checking are represented in Figure \ref{sub}. For subsumption of concepts or roles, we find that LLMs achieve promising results in most cases. However, for instance checking, none of the logical implications can be inferred by LLMs for UOBM1 and UOBM2, even though LLMs achieve good performances for the other three ontologies.  This is because the task of subsumption of concepts or roles only requires the input of the TBox which is usually relatively small, while instance checking requires an ontology that includes both the TBox and the ABox where sometimes the ABox can be quite large. We input the TBox and ABox at one prompt and the size of  UOBM1 and UOBM2 exceeds the maximum size limit that the selected LLMs can handle. Overall, LLMs perform well in these two tasks when the input ontology is relatively small. More specifically, we analyze the following questions:

\textbf{\textit{How do the size of the ontology and the scale of LLMs affect the  understanding of the ontology?}} 
The experimental results show that the larger  the ontology is, the worse the understanding of LLMs is. For small ontologies, LLMs can achieve almost 100\% performance. However, when the size of the ontology exceeds a certain threshold, the performance of LLMs drops to nearly 0\%. Similarly, the larger  the scale of the LLM is,  the better its capacity to understand ontologies is. For instance, the scale of LLaMA3-8B is much smaller than that of GPT-3.5 and GPT-4o, so its performances on several ontologies are significantly worse.

\textbf{\textit{Can LLMs understand the transitivity rules and efficiently apply them in reasoning?}} For subsumption of concepts or roles and the smaller three ontologies UOBM3, UOBM4, UOBM5 in instance checking in Figure \ref{sub}, GPT4o can deduce all the implications and GPT3.5, LLaMA3-7b can both deduce most of the logical implications, indicating that LLMs can efficiently perform instance checking and subsumption of concepts or roles when the ontology is not that large.

However, this does not mean that LLMs truly understand and correctly use every transitivity rule because: (1) The used transitivity rules for those logical implications only cover a small part of all the transitivity rules; (2) LLMs may have potential hallucinations about transitivity rules. Thus we conduct a case study. We build five handcrafted DL-Lite ontologies with logical implications for this use where each logical implication can be deduced by certain kind of transitivity rule and the examples cover all the introduced transitivity rules, as shown in Table \ref{transitivity}.   We apply the above prompts but add ``Give reasons or inferring process for each answer.'' to the end of task definition (\textit{T}). Figure \ref{transitivity-results} shows the results. LLMs perform well in case 2, case 4 and case 5, but perform poorly in case 1 and case 3, because most logical implications in  case 1 and case 3 can only be deduced by TBox NI transitivity, and those in other cases can be deduced by TBox PI transitivity or ABox transitivity. LLMs fail to understand TBox NI transitivity rules well, and instructions or examples have limited effect. We also find  LLMs give incorrect explanations to logical implications which can only be deduced by certain TBox NI transitivity rules, indicating that LLMs have hallucinations about TBox NI transitivity rules or possess some incorrect knowledge about TBox NI transitivity.

\begin{table*}[ht]
\centering
\scalebox{0.8}{
\begin{tabular}{cccccccc}
\toprule
Data Sources& \#ax. & \#as. & \#inv. & \#fun. & \#inv. fun. & \#impli.inv. & \#impli.fun.  \\
\midrule
Academic Hierarchy & 36 & 120 & 6 & 3 & 1 & 12 & 12 \\
E-Commerce  System & 32 & 51 & 4 & 2 & 1 & 8 & 6 \\
Library System & 22 & 70 & 3 & 0 & 3 & 7 & 6 \\
Social Network Relations & 29 & 102 & 3 & 1 & 4 & 5 & 3 \\
Medical Medical Relationships & 16 & 21 & 3 & 1 & 0 & 13 & 4 \\
\bottomrule
\end{tabular}}
\caption{Statistics about data sources for property characteristics probing.  \# denotes ``the number of '', and ax., as., inv., fun., inv. fun., impli., impli. fun. denote class axioms, class assertions, inverse roles, functional roles, inverse functional roles, logical implications for inverse roles, logical implication for functional roles.}
\label{property-data}
\end{table*}

\subsubsection{Property Characteristics Probing} \label{C}
Property characteristics, such as \textit{symmetric property},  \textit{transitive property}, \textit{functional property} and \textit{inverse functional property}, play a significant role in a DL ontology. Some studies have shown evidence that the LLMs have limited knowledge of some property characteristics without external knowledge or instructions such as inverse role property (called "reversal curse" in \cite{reverse}) and  property inheritance \cite{concept-aware}. In this work, especially, we focus on two important property characteristics in DL-Lite, inverse role property and (inverse) functional role property.
We set property characteristics probing tasks: \\
$\bullet$ inverse role probing: Given an ontology $O$, a role $R$, its inverse role $P=R^{-}$, and two constants $a$ and $b$ which satisfy $O \mid=R(a, b)$, verify whether $O \mid=P(b, a)$. \\
$\bullet$ (inverse) functional role probing: Given an ontology $O$, a functional role (funct $R$) (an inverse functional role (funct $R^{-}$)), and three constants $a, b$ and $c$ which satisfies $O \mid=R(a, b)$ and $O \mid=R(a, c)$ (resp. $O \mid=$ $R(b, a)$ and $O \mid=R(c, a))$, verify whether $b \equiv c$.

\textbf{Datasets.}  We obtain the DL-Lite datasets by extracting and processing existing DL ontologies, namely, Academic Hierarchy (from the University Ontology Benchmark \cite{uobm}), E-Commerce System (from the GoodRelations Ontology \cite{goodrelations}), Library System (from the Dublin Core Metadata \cite{dublin}), Social Network Relations (from FOAF, Friend of a Friend \cite{foaf}) and Medical Relationships (from SNOMED CT \cite{ct}). Table \ref{property-data} shows statistics about data sources for property characteristics probing.

For inverse role property probing, we select inverse roles in the ontologies and use them to build logical implications. For example, if $WorksIn$ and $WorksIn^{-}$ exists, we add $Employs$, $Employs$$\sqsubseteq$$WorksIn^{-}$, $WorksIn^{-}$$\sqsubseteq$$Employs$ to the ontology. If $WorksIn(a, b)$ exists in the ontology, we build the logical implication $Employs(b, a)$. For (inverse) functional role property probing, similarly we select functional roles and build logical implications. For example, if $(funct \ BelongsTo)$ and $BelongsTo(a, b)$ hold, we then add $BelongsTo(a, x)$ to the ontology and build the logical implication $x \equiv b$. Statistical details are covered in Appendix \ref{AF}.

\textbf{Experimental setup.} The prompt is almost the same to  prompt-\textit{NI} in instance checking. We add ``Give reasons or inferring process." to the end of the task definition.  We use GPT4o and the same metric in instance checking for evaluation.

\textbf{Results analysis.} The results in Figure \ref{property} show that LLMs can deduce most of the logical implications. LLMs give reasonable explanations of the deducing process such as ``Since  BelongsTo(Product1, Category1) is given  and  BelongsTo is the inverse of Owns, hence Owns(Category1, Product1) can be deduced'' and ``Given: 
WorksAt(DrBrown,RegionalHospital) and
WorksAt(DrBrown,x3). Since WorksAt is a functional property, DrBrown can only work at one hospital. Thus, x3 must be RegionalHospital to satisfy the functional constraint''.  LLMs have the potential to understand such logical constraints in DL ontologies, indicating the promising prospects to utlize ontologies to enhance LLMs' inference capacity such as in the scene of ``reversal curse'' \cite{reverse}.

\subsubsection{Query Answering} \label{D}
Query answering over an ontology involves retrieving information that satisfies a given query based on this ontology \cite{dl-lite}. 

\textbf{Datasets.} We use the Lehigh University Benchmark (LUBM) \cite{lubm} with the given TBox, ABox example and  14 test queries\footnote{https://swat.cse.lehigh.edu/projects/lubm/}. 

\textbf{Experimental setup.} We use GPT4o for evaluation. Similar to prompt-\textit{NI} in instance checking, the prompt includes task description (\textit{T}), input ontology (\textit{O}) and the query (\textit{Q}). Because LLMs can't handle large-scale ABox at one time as shown in Section \ref{A}, we cut the ontology into 10 parts and  input them in turn.

\textbf{Results analysis.} Test results show that GPT4o fails to give a totally correct answer for each query. For Q3, Q8, Q12, Q13 and Q14, GPT4o can only answer a very small part of all the expected answers. For other queries, GPT4o has hallucinations and answer incorrect answers. For
example, the expect answer is {Student0, Student3, Student9} but LLM answers {Student2, Student4}. LLMs can't memorize and understand large scale factual knowledge and fail to perform query answering well practically.

\subsubsection{Ontology Satisfiability Checking} \label{E}
Ontology satisfiability checking is to verify the logical consistency of an ontology by ensuring the existence of at least one model that satisfies its axioms. This process is closely related to the semantic relationships within the ontology because a consistent, semantically meaningful ontology is more likely to be satisfiable and able to provide an accurate representation of the intended domain.

\textbf{Datasets.}  We build inconsistent DL-Lite ontologies by generating minimal inconsistent subsets (MISs) \cite{mis} of existing inconsistent ontologies from \cite{inconsistent}. We choose economy-Inc. and Maa-edas-iasted in \cite{inconsistent} to generate MISs, because the expressivity of their MISs is close to DL-Lite. We select 29 MISs of economy-Inc.  and 38 MISs of Maa-edas-iasted. For each MIS, we randomly delete an axiom to obtain the corresponding consistent ontology.

\textbf{Experimental setup.} The experimental settings are similar to those in syntax checking. We use the prompt-\textit{NI} including task definition ($T$) and  ontology ($O$). We cover prompts in Appendix \ref{AC}.

\begin{table}
\centering
\scalebox{0.57}{
\begin{tabular}{c|ccc|ccc}
\hline Datasets & \multicolumn{3}{c|}{economy-inc.} & \multicolumn{3}{c}{MaasMatch.}  \\ \hline
 Metric  & Precision & Recall & F1-Score  & Precision & Recall & F1-Score \\ \hline
GPT3.5--\textit{NI} & 100 & 93.1 & 96.4 & 57.6 & 100 & 73.1  \\ 
GPT4o--\textit{NI}   & 100 &  89.7 & 94.5  &  63.0 & 76.3 & 69.0 \\ 
LLaMA3-8b--\textit{NI} & 81.0 & 58.6 & 68.0 & 55.3 & 55.3 &  55.3 \\ 
\hline
\end{tabular}}
\caption{Performances of LLMs in ontology satisfiability checking (\%).}
 \label{incons}
\end{table}

\textbf{Results analysis.} From Table \ref{incons}, we observe that LLMs perform well on economiy-inc., and relatively poor on Maa-edas-iasted, since Maa-edas-iasted is more complex and has more constructors. Overall, LLMs can detect logical inconsistencies in DL-Lite ontologies to some degree. However, this  capacity is limited for more complex inconsistent DL ontologies.

\section{Conclusion}
We have empirically investigated whether LLMs can understand DL-Lite ontologies.  Extensive experimental results demonstrated the effectiveness and limitations of LLMs in understanding the syntax and semantics of DL-Lite ontologies. For instance, LLMs possess the ability to understand formal syntax and semantics of concepts, roles and property characteristics.  However, LLMs still struggle with understanding TBox NI transitivity rules and handling ontologies with large scale ABoxes.

As future works, we will consider exploring the ability of LLMs to understand ontologies in other lightweight ontology languages, such as $\mathcal{EL}$, and to understand ontologies in intractable ontology languages, such as $\mathcal{ALC}$ and $\mathcal{SHOIQ}$.
 
\section*{Limitations}
This work is an empirical study on LLMs' capacity of understanding DL-Lite ontologies, and it has several
limitations. Firstly, the size and diversity are limited due to the data sources and costs of LLMs. Secondly, there are various kinds of DLs and we just choose DL-Lite for evaluation. We  thus encourage future work to conduct investigations for more DLs. Finally, it still remains unexplored  how to improve LLMs' understanding capacity for TBox NI transitivity and large-scale ABox.

\section*{Acknowledgements}
This work is partially supported by National Nature Science Foundation of China under No. U21A20488 and by the project "Key Laboratory of rich-media Digital Publishing Content Organization and Knowledge Service Open Fund-Research on Knowledge-enhanced Training Techinques of Large Language Model"No. ZD2024-04/01. We thank the Big Data Computing Center of Southeast University for providing the facility support on the numerical calculations in this paper.

\bibliography{acl}

\begin{thebibliography}{43}
\providecommand{\natexlab}[1]{#1}

\bibitem[{Bandrowski et~al.(2016)Bandrowski, Brinkman, Brochhausen, Brush, Bug, Chibucos, Clancy, Courtot, Derom, Dumontier et~al.}]{obi}
Anita Bandrowski, Ryan Brinkman, Mathias Brochhausen, Matthew~H Brush, Bill Bug, Marcus~C Chibucos, Kevin Clancy, M{\'e}lanie Courtot, Dirk Derom, Michel Dumontier, et~al. 2016.
\newblock The ontology for biomedical investigations.
\newblock \emph{PloS one}, 11(4):e0154556.

\bibitem[{Bao et~al.(2024)Bao, Zhang, Yang, Wang, and Zhang}]{logic2}
Guangsheng Bao, Hongbo Zhang, Linyi Yang, Cunxiang Wang, and Yue Zhang. 2024.
\newblock Llms with chain-of-thought are non-causal reasoners.
\newblock \emph{arXiv preprint arXiv:2402.16048}.

\bibitem[{Berglund et~al.(2023)Berglund, Tong, Kaufmann, Balesni, Stickland, Korbak, and Evans}]{reverse}
Lukas Berglund, Meg Tong, Max Kaufmann, Mikita Balesni, Asa~Cooper Stickland, Tomasz Korbak, and Owain Evans. 2023.
\newblock The reversal curse: Llms trained on" a is b" fail to learn" b is a".
\newblock \emph{arXiv preprint arXiv:2309.12288}.

\bibitem[{Bouaud et~al.(1995)Bouaud, Bachimont, Charlet, and Zweigenbaum}]{exin1}
Jacques Bouaud, Bruno Bachimont, Jean Charlet, and Pierre Zweigenbaum. 1995.
\newblock Methodological principles for structuring an “ontology”.
\newblock In \emph{Proceedings of the IJCAI’95 Workshop on “Basic Ontological Issues in Knowledge Sharing}, pages 19--25.

\bibitem[{Brown et~al.(2020{\natexlab{a}})Brown, Mann, Ryder, Subbiah, Kaplan, Dhariwal, Neelakantan, Shyam, Sastry, Askell et~al.}]{gpt}
Tom Brown, Benjamin Mann, Nick Ryder, Melanie Subbiah, Jared~D Kaplan, Prafulla Dhariwal, Arvind Neelakantan, Pranav Shyam, Girish Sastry, Amanda Askell, et~al. 2020{\natexlab{a}}.
\newblock Language models are few-shot learners.
\newblock \emph{Advances in neural information processing systems}, 33:1877--1901.

\bibitem[{Brown et~al.(2020{\natexlab{b}})Brown, Mann, Ryder, Subbiah, Kaplan, Dhariwal, Neelakantan, Shyam, Sastry, Askell et~al.}]{gpt3.5-1}
Tom Brown, Benjamin Mann, Nick Ryder, Melanie Subbiah, Jared~D Kaplan, Prafulla Dhariwal, Arvind Neelakantan, Pranav Shyam, Girish Sastry, Amanda Askell, et~al. 2020{\natexlab{b}}.
\newblock Language models are few-shot learners.
\newblock \emph{Advances in neural information processing systems}, 33:1877--1901.

\bibitem[{Calvanese et~al.(2009)Calvanese, De~Giacomo, Lembo, Lenzerini, Poggi, Rodriguez-Muro, and Rosati}]{dl-lite-onto}
Diego Calvanese, Giuseppe De~Giacomo, Domenico Lembo, Maurizio Lenzerini, Antonella Poggi, Mariano Rodriguez-Muro, and Riccardo Rosati. 2009.
\newblock Ontologies and databases: The dl-lite approach.
\newblock In \emph{Reasoning Web International Summer School}, pages 255--356. Springer.

\bibitem[{Calvanese et~al.(2007)Calvanese, De~Giacomo, Lembo, Lenzerini, and Rosati}]{dl-lite}
Diego Calvanese, Giuseppe De~Giacomo, Domenico Lembo, Maurizio Lenzerini, and Riccardo Rosati. 2007.
\newblock Tractable reasoning and efficient query answering in description logics: The dl-lite family.
\newblock \emph{Journal of Automated reasoning}, 39:385--429.

\bibitem[{Chen et~al.(2023)Chen, Ma, Song, Cao, Zhang, and Li}]{logic5}
Meiqi Chen, Yubo Ma, Kaitao Song, Yixin Cao, Yan Zhang, and Dongsheng Li. 2023.
\newblock Learning to teach large language models logical reasoning.
\newblock \emph{arXiv preprint arXiv:2310.09158}.

\bibitem[{Consortium(2004)}]{go}
Gene~Ontology Consortium. 2004.
\newblock The gene ontology (go) database and informatics resource.
\newblock \emph{Nucleic acids research}, 32(suppl\_1):D258--D261.

\bibitem[{El-Sappagh et~al.(2018)El-Sappagh, Franda, Ali, and Kwak}]{ct}
Shaker El-Sappagh, Francesco Franda, Farman Ali, and Kyung-Sup Kwak. 2018.
\newblock Snomed ct standard ontology based on the ontology for general medical science.
\newblock \emph{BMC medical informatics and decision making}, 18:1--19.

\bibitem[{Feng et~al.(2023)Feng, Zhang, and Fei}]{knowledge-solver}
Chao Feng, Xinyu Zhang, and Zichu Fei. 2023.
\newblock Knowledge solver: Teaching llms to search for domain knowledge from knowledge graphs.
\newblock \emph{arXiv preprint arXiv:2309.03118}.

\bibitem[{Fill et~al.(2023)Fill, Fettke, and K{\"o}pke}]{LLM4EL}
Hans-Georg Fill, Peter Fettke, and Julius K{\"o}pke. 2023.
\newblock Conceptual modeling and large language models: impressions from first experiments with chatgpt.
\newblock \emph{Enterprise Modelling and Information Systems Architectures (EMISAJ)}, 18:1--15.

\bibitem[{Formica(2006)}]{exin3}
Anna Formica. 2006.
\newblock Ontology-based concept similarity in formal concept analysis.
\newblock \emph{Information sciences}, 176(18):2624--2641.

\bibitem[{Glimm et~al.(2014)Glimm, Horrocks, Motik, Stoilos, and Wang}]{hermit}
Birte Glimm, Ian Horrocks, Boris Motik, Giorgos Stoilos, and Zhe Wang. 2014.
\newblock Hermit: an owl 2 reasoner.
\newblock \emph{Journal of automated reasoning}, 53:245--269.

\bibitem[{Golbeck and Rothstein(2008)}]{foaf}
Jennifer Golbeck and Matthew Rothstein. 2008.
\newblock Linking social networks on the web with foaf: A semantic web case study.
\newblock In \emph{AAAI}, volume~8, pages 1138--1143.

\bibitem[{Guo et~al.(2005)Guo, Pan, and Heflin}]{lubm}
Yuanbo Guo, Zhengxiang Pan, and Jeff Heflin. 2005.
\newblock Lubm: A benchmark for owl knowledge base systems.
\newblock \emph{Journal of Web Semantics}, 3(2-3):158--182.

\bibitem[{Heinzerling and Inui(2021)}]{lmkb}
Benjamin Heinzerling and Kentaro Inui. 2021.
\newblock \href {https://doi.org/10.18653/v1/2021.eacl-main.153} {Language models as knowledge bases: On entity representations, storage capacity, and paraphrased queries}.
\newblock In \emph{Proceedings of the 16th Conference of the European Chapter of the Association for Computational Linguistics: Main Volume}, pages 1772--1791, Online. Association for Computational Linguistics.

\bibitem[{Hepp(2008)}]{goodrelations}
Martin Hepp. 2008.
\newblock Goodrelations: An ontology for describing products and services offers on the web.
\newblock In \emph{Knowledge Engineering: Practice and Patterns: 16th International Conference, EKAW 2008, Acitrezza, Italy, September 29-October 2, 2008. Proceedings 16}, pages 329--346. Springer.

\bibitem[{Hunter et~al.(2008)Hunter, Konieczny et~al.}]{mis}
Anthony Hunter, S{\'e}bastien Konieczny, et~al. 2008.
\newblock Measuring inconsistency through minimal inconsistent sets.
\newblock \emph{KR}, 8(358-366):42.

\bibitem[{Ji et~al.(2014)Ji, Gao, Huang, and Zhu}]{inconsistent}
Qiu Ji, Zhiqiang Gao, Zhisheng Huang, and Man Zhu. 2014.
\newblock Measuring effectiveness of ontology debugging systems.
\newblock \emph{Knowledge-Based Systems}, 71:169--186.

\bibitem[{Keet et~al.(2008)Keet, Alberts, Gerber, and Chimamiwa}]{adolena}
C~Maria Keet, Ronell Alberts, Aurona Gerber, and Gibson Chimamiwa. 2008.
\newblock Enhancing web portals with ontology-based data access: The case study of south africa's accessibility portal for people with disabilities.
\newblock In \emph{OWLED}, volume 432.

\bibitem[{Luo et~al.(2023)Luo, Kumbhar, Parmar, Varshney, Banerjee, Aditya, Baral et~al.}]{logic3}
Man Luo, Shrinidhi Kumbhar, Mihir Parmar, Neeraj Varshney, Pratyay Banerjee, Somak Aditya, Chitta Baral, et~al. 2023.
\newblock Towards logiglue: A brief survey and a benchmark for analyzing logical reasoning capabilities of language models.
\newblock \emph{arXiv preprint arXiv:2310.00836}.

\bibitem[{Ma et~al.(2006)Ma, Yang, Qiu, Xie, Pan, and Liu}]{uobm}
Li~Ma, Yang Yang, Zhaoming Qiu, Guotong Xie, Yue Pan, and Shengping Liu. 2006.
\newblock Towards a complete owl ontology benchmark.
\newblock In \emph{The Semantic Web: Research and Applications: 3rd European Semantic Web Conference, ESWC 2006 Budva, Montenegro, June 11-14, 2006 Proceedings 3}, pages 125--139. Springer.

\bibitem[{Mateiu and Groza(2023)}]{LLM4OWL}
Patricia Mateiu and Adrian Groza. 2023.
\newblock \href {https://api.semanticscholar.org/CorpusID:260333920} {Ontology engineering with large language models}.
\newblock \emph{2023 25th International Symposium on Symbolic and Numeric Algorithms for Scientific Computing (SYNASC)}, pages 226--229.

\bibitem[{Motik and Sattler(2006)}]{data2}
Boris Motik and Ulrike Sattler. 2006.
\newblock A comparison of reasoning techniques for querying large description logic aboxes.
\newblock In \emph{International Conference on Logic for Programming Artificial Intelligence and Reasoning}, pages 227--241. Springer.

\bibitem[{Mruthyunjaya et~al.(2023)Mruthyunjaya, Pezeshkpour, Hruschka, and Bhutani}]{llm4skg}
Vishwas Mruthyunjaya, Pouya Pezeshkpour, Estevam Hruschka, and Nikita Bhutani. 2023.
\newblock Rethinking language models as symbolic knowledge graphs.
\newblock \emph{arXiv preprint arXiv:2308.13676}.

\bibitem[{Nagyp{\'a}l et~al.(2005)Nagyp{\'a}l, Deswarte, and Oosthoek}]{vicodi}
G{\'a}bor Nagyp{\'a}l, Richard Deswarte, and Jan Oosthoek. 2005.
\newblock Applying the semantic web: The vicodi experience in creating visual contextualization for history.
\newblock \emph{Literary and Linguistic Computing}, 20(3):327--349.

\bibitem[{OpenAI(2023)}]{gpt4}
R~OpenAI. 2023.
\newblock Gpt-4 technical report. arxiv 2303.08774.
\newblock \emph{View in Article}, 2(5).

\bibitem[{Pan et~al.(2023)Pan, Albalak, Wang, and Wang}]{logic4}
Liangming Pan, Alon Albalak, Xinyi Wang, and William Wang. 2023.
\newblock \href {https://doi.org/10.18653/v1/2023.findings-emnlp.248} {Logic-{LM}: Empowering large language models with symbolic solvers for faithful logical reasoning}.
\newblock In \emph{Findings of the Association for Computational Linguistics: EMNLP 2023}, pages 3806--3824, Singapore. Association for Computational Linguistics.

\bibitem[{P{\'e}rez-Urbina et~al.(2009)P{\'e}rez-Urbina, Motik, and Horrocks}]{data1}
H{\'e}ctor P{\'e}rez-Urbina, Boris Motik, and Ian Horrocks. 2009.
\newblock A comparison of query rewriting techniques for dl-lite.
\newblock \emph{Description Logics}, 477:29.

\bibitem[{Raimond and Sandler(2012)}]{music}
Yves Raimond and Mark Sandler. 2012.
\newblock Evaluation of the music ontology framework.
\newblock In \emph{Extended Semantic Web Conference}, pages 255--269. Springer.

\bibitem[{Rodriguez-Muro et~al.(2008)Rodriguez-Muro, Lubyte, and Calvanese}]{stockexechange}
Mariano Rodriguez-Muro, Lina Lubyte, and Diego Calvanese. 2008.
\newblock Realizing ontology based data access: A plug-in for prot{\'e}g{\'e}.
\newblock In \emph{2008 IEEE 24th International Conference on Data Engineering Workshop}, pages 286--289. IEEE.

\bibitem[{Rosse and Mejino~Jr(2008)}]{fma}
Cornelius Rosse and Jos{\'e}~LV Mejino~Jr. 2008.
\newblock The foundational model of anatomy ontology.
\newblock In \emph{Anatomy ontologies for bioinformatics: principles and practice}, pages 59--117. Springer.

\bibitem[{Shani et~al.(2023)Shani, Vreeken, and Shahaf}]{concept-aware}
Chen Shani, Jilles Vreeken, and Dafna Shahaf. 2023.
\newblock \href {https://doi.org/10.18653/v1/2023.findings-emnlp.877} {Towards concept-aware large language models}.
\newblock In \emph{Findings of the Association for Computational Linguistics: EMNLP 2023}, pages 13158--13170, Singapore. Association for Computational Linguistics.

\bibitem[{Touvron et~al.(2023)Touvron, Lavril, Izacard, Martinet, Lachaux, Lacroix, Rozi{\`e}re, Goyal, Hambro, Azhar et~al.}]{llama}
Hugo Touvron, Thibaut Lavril, Gautier Izacard, Xavier Martinet, Marie-Anne Lachaux, Timoth{\'e}e Lacroix, Baptiste Rozi{\`e}re, Naman Goyal, Eric Hambro, Faisal Azhar, et~al. 2023.
\newblock Llama: Open and efficient foundation language models.
\newblock \emph{arXiv preprint arXiv:2302.13971}.

\bibitem[{Tzitzikas et~al.(2016)Tzitzikas, Allocca, Bekiari, Marketakis, Fafalios, Doerr, Minadakis, Patkos, and Candela}]{marine}
Yannis Tzitzikas, Carlo Allocca, Chryssoula Bekiari, Yannis Marketakis, Pavlos Fafalios, Martin Doerr, Nikos Minadakis, Theodore Patkos, and Leonardo Candela. 2016.
\newblock Unifying heterogeneous and distributed information about marine species through the top level ontology marinetlo.
\newblock \emph{Program}, 50(1):16--40.

\bibitem[{Wang et~al.(2024{\natexlab{a}})Wang, Qi, Chen, and Wu}]{eike}
Keyu Wang, Guilin Qi, Jiaoyan Chen, and Tianxing Wu. 2024{\natexlab{a}}.
\newblock Embedding ontologies via incoprorating extensional and intensional knowledge.
\newblock \emph{arXiv preprint arXiv:2402.01677}.

\bibitem[{Wang et~al.(2024{\natexlab{b}})Wang, Wei, Choi, and Ren}]{logic1}
Siyuan Wang, Zhongyu Wei, Yejin Choi, and Xiang Ren. 2024{\natexlab{b}}.
\newblock Can llms reason with rules? logic scaffolding for stress-testing and improving llms.
\newblock \emph{arXiv preprint arXiv:2402.11442}.

\bibitem[{Weibel et~al.(1998)Weibel, Kunze, Lagoze, and Wolf}]{dublin}
Stuart Weibel, John Kunze, Carl Lagoze, and Misha Wolf. 1998.
\newblock Dublin core metadata for resource discovery.
\newblock Technical report.

\bibitem[{Woods(1975)}]{exin2}
William~A Woods. 1975.
\newblock What's in a link: Foundations for semantic networks.
\newblock In \emph{Representation and understanding}, pages 35--82. Elsevier.

\bibitem[{Yang et~al.(2023)Yang, Xiong, Payani, Shareghi, and Fekri}]{LLM4FOL}
Yuan Yang, Siheng Xiong, Ali Payani, Ehsan Shareghi, and Faramarz Fekri. 2023.
\newblock Harnessing the power of large language models for natural language to first-order logic translation.
\newblock \emph{arXiv preprint arXiv:2305.15541}.

\bibitem[{Zheng et~al.(2024)Zheng, Qiu, Shi, and Ma}]{xiaoming}
Junhao Zheng, Shengjie Qiu, Chengming Shi, and Qianli Ma. 2024.
\newblock Towards lifelong learning of large language models: A survey.
\newblock \emph{arXiv preprint arXiv:2406.06391}.

\end{thebibliography}

\clearpage

\appendix


\section{DL-Lite Transitivity Rules} \label{AA}
\noindent \fbox{\small{\parbox{0.9\textwidth}{%
\textbf{TBox PI transitivity rules:}\\
$\alpha=C_1 \sqsubseteq C_2,   \beta=C_2 \sqsubseteq C_3   \rightarrow  \beta_{\text{new}}=C_1 \sqsubseteq C_3$\\
$\alpha=R_1 \sqsubseteq R_2 ,  \beta=R_2 \sqsubseteq R_3   \rightarrow  \beta_{\text{new}}=R_1 \sqsubseteq R_3$\\ \\ 
\textbf{TBox NI transitivity rules:}\\
$ \alpha=C_1 \sqsubseteq C_2 ,  \beta=C_2 \sqsubseteq \neg C_3  \rightarrow \beta_{\text {new }}=C_1 \sqsubseteq \neg C_3 $\\
$\alpha=C_1 \sqsubseteq C_2,  \beta=C_3 \sqsubseteq \neg C_2 \rightarrow \beta_{\text {new}}=C_1 \sqsubseteq \neg C_3 \\
 \alpha=R_1 \sqsubseteq R_2, \beta=\exists R_2 \sqsubseteq \neg C \rightarrow \beta_{\text {new }}=\exists R_1 \sqsubseteq \neg C $\\
$\alpha=R_1 \sqsubseteq R_2, \beta=C \sqsubseteq \neg \exists R_2 \rightarrow \beta_{\text {new }}=\exists R_1 \sqsubseteq \neg C $\\
$ \alpha=R_1 \sqsubseteq R_2, \beta=C \sqsubseteq \neg \exists R_2^{-} \rightarrow \beta_{\text {new }}=\exists R_1^{-} \sqsubseteq \neg C $\\
$ \alpha=R_1 \sqsubseteq R_2, \beta=\exists R_2^{-} \sqsubseteq \neg C \rightarrow \beta_{\text {new }}=\exists R_1^{-} \sqsubseteq \neg C $\\
$ \alpha=R_1 \sqsubseteq R_2, \beta=R_2 \sqsubseteq \neg R_3  \rightarrow \beta_{\text {new }}=R_1 \sqsubseteq \neg R_3 $\\
$ \alpha=R_1 \sqsubseteq R_2, \beta=R_3 \sqsubseteq \neg R_2  \rightarrow \beta_{\text {new }}=R_1 \sqsubseteq \neg R_3 $\\
$ \alpha=R \sqsubseteq \neg R \rightarrow \beta_{\text {new}_{1}}=\exists R \sqsubseteq \neg \exists R, \beta_{\text {new}_{2}}=\exists R^{-} \sqsubseteq \neg \exists R^{-} $ \\
$ \alpha=\exists R \sqsubseteq \neg \exists R \rightarrow \beta_{\text {new}_{1}}=R \sqsubseteq \neg R, \beta_{\text {new}_{2}}=\exists R^{-} \sqsubseteq \neg \exists R^{-} $ \\
$ \alpha=\exists R^{-} \sqsubseteq \neg \exists R^{-}  \rightarrow \beta_{\text {new}_{1}}=R \sqsubseteq \neg R, \beta_{\text {new}_{2}}=\exists R \sqsubseteq \neg \exists R $ 
\\ \\ 
\textbf{ABox transitivity rules:} \\ 
$ \alpha=C_{1} \sqsubseteq C_{2}, \beta=C_{1}(a)  \rightarrow  \beta_{\text {new }}=C_{1}(a) $\\
$\alpha=C \sqsubseteq \exists R , \beta=C(a)  \rightarrow  \beta_{\text {new }}=R\left(a, a_{\text {new}}\right) $\\
$ \alpha=\exists R \sqsubseteq C, \beta=R\left(a, a^{\prime}\right)  \rightarrow  \beta_{\text {new }}=C(a) $\\
$\alpha=\exists R_{1} \sqsubseteq \exists R_{2}, \beta=R_{1}\left(a, a^{\prime}\right)  \rightarrow  \beta_{\text {new }}=R_{2}\left(a, a_{\text {new }}\right)$ \\
$ \alpha=R_{1} \sqsubseteq R_{2},  \beta=R_{1}\left(a, a^{\prime}\right)  \rightarrow  \beta_{\text {new }}=R_{2}\left(a, a^{\prime}\right)$ 
}}} \\ 

We refer to Section 3.1 in \cite{dl-lite} for detailed illustrations and examples about these transitivity roles.  \\ \\

\section{Typical DL-Lite Syntax Errors} \label{AB}

\scalebox{1.0}{
\begin{tabular}{ccc}
\toprule

 & \textbf{Common Syntax Errors in DL-Lite} & \textbf{Examples} \\
\midrule
\multirow{3}{*}{Invalid inverse } & Inverse operator on a concept & $ Professor^{-}$  \\
  & Misplaced inverse operator  & $^{-}TeachesTo$  \\
 & Inverse operator on a quantifier  & $ \exists^{-}$ \\
\midrule
\multirow{4}{*}{Invalid quantifiers} & Misplaced quantifiers & $TeachesTo \exists$ \\
  & Quantifiers with concept following     & $\exists Professor$ \\
 & Quantifiers missing role following    & $ \exists  $\\
  & Redundant multiple quantifiers    & $ \exists \exists TeachesTo $\\
\midrule
\multirow{2}{*}{Invalid negation} & Misplaced negation operator & $ Professor\neg $ \\
  & Negation without anything following & $\neg$ \\
\midrule
\multirow{5}{*}{Invalid conjunction} & Conjoining incomplete concepts & $ Professor \sqcap $  \\
  & Conjoining a concept with a role & $ Professor \sqcap TeachesTo $ \\
 & Conjoining roles directly & $ TeachesTo \sqcap  HasTutor $\\
  & Missing conjunction operator & $Professor \exists TeachesTo$\\
   & Misplaced conjunction operator & $\sqcap Professor \exists TeachesTo$\\
\bottomrule
\end{tabular}}

\clearpage

\section{Prompts And Answer Examples} \label{AC}
All symbols and constructors in the prompts can be input into LLMs, but only one kind of font can be input into LLMs (Colors and italics are only for display convenience). \\

\setlength{\parindent}{0pt} \textbf{Prompt-\textit{NI} for syntax checking:} \\  \vspace{-3mm}
\hrule 
\vspace{1mm}
\setlength{\parindent}{0pt} \textcolor{blue}{Task Description}: \\
There are some DL-Lite axioms, and your task is to determine whether the syntax of each of these axioms is correct. \\ \\
\setlength{\parindent}{0pt} \textcolor{blue}{Given DL-Lite Axioms: } \\
MaterialEntity $\sqsubseteq$ $\neg$PhysicalObject \\
$\exists$hasPerformer$\neg$ $\sqsubseteq$ MusicalExpression \\
Investigation $\sqsubseteq$ $\exists$hasPart \\ 
Protocol $\sqsubseteq$ $\neg$Investigation$\neg$ \\
\textcolor{gray}{(··· more context here ···)} \\  \\
\setlength{\parindent}{0pt} \textcolor{blue}{Answer: }  
\vspace{1mm}
\hrule 

\vspace{1cm}

\setlength{\parindent}{0pt} \textbf{Prompt-\textit{WI} for syntax checking:} \\  \vspace{-3mm}
\hrule 
\vspace{1mm}
\textcolor{blue}{Task Description}: \\
There are some DL-Lite axioms, and your task is to determine whether the syntax of each of these axioms is correct. \\
DL-Lite{$_{core}$} concepts and roles are defined as follows: \\
$ B ::= A \ | \ \exists R \ | \ \exists R^{-}   \quad  R ::= P \ | \ P^{-}\\ \quad C ::= B \ | \ \neg B \ | \ C_1 \sqcap C_2 \quad E ::= R \ | \ \neg R $ \\
where $A$ denotes an atomic concept, $P$ denotes an atomic role, and $P^-$ denotes the inverse of the atomic role $P$. We call $B,R,C,E$ a basic concept, a basic role, a general concept and a general role respectively. \\
A DL-Lite$_{\textit{core}}$ ontology $\mathcal{O}=\langle\mathcal{T}, \mathcal{A}\rangle$ consists of a TBox $\mathcal{T}$ and an ABox $\mathcal{A}$.  $\mathcal{T}$ is formed by a finite set of concept inclusion assertions of the form $B \sqsubseteq \mathrm{C}$. $\mathcal{A}$ is formed by a finite set of membership assertions on atomic concepts and on atomic roles, of the form $A(a)$ and $P(a,b)$. DL-Lite$_{\mathcal{R}}$ extends DL-Lite$_{\text {core}}$ with role inclusion assertions of the form $R  \sqsubseteq E$ and DL-Lite$_{\mathcal{F}}$ extends DL-Lite$_{\textit{core}}$ with functionality on roles or on their inverses of the form (Funct $R$). \\ \\
\setlength{\parindent}{0pt} \textcolor{blue}{Given DL-Lite Axioms: } \\
MaterialEntity $\sqsubseteq$ $\neg$PhysicalObject \\
$\exists$hasPerformer$\neg$ $\sqsubseteq$ MusicalExpression \\
Investigation $\sqsubseteq$ $\exists$hasPart \\ 
Protocol $\sqsubseteq$ $\neg$Investigation$\neg$ \\
\textcolor{gray}{(··· more context here ···)} \\ \\
\setlength{\parindent}{0pt} \textcolor{blue}{Answer: } 
\vspace{1mm}
\hrule 

\vspace{1cm}

\setlength{\parindent}{0pt} \textbf{Prompt-\textit{WIE} for syntax checking:} \\  \vspace{-3mm}
\hrule 
\vspace{1mm}
\textcolor{blue}{Task Description}: \\
There are some DL-Lite axioms, and your task is to determine whether the syntax of each of these axioms is correct. \\
DL-Lite{$_{core}$} concepts and roles are defined as follows: \\
$ B ::= A \ | \ \exists R \ | \ \exists R^{-}   \quad  R ::= P \ | \ P^{-}\\ \quad C ::= B \ | \ \neg B \ | \ C_1 \sqcap C_2 \quad E ::= R \ | \ \neg R $ \\
where $A$ denotes an atomic concept, $P$ denotes an atomic role, and $P^-$ denotes the inverse of the atomic role $P$. We call $B,R,C,E$ a basic concept, a basic role, a general concept and a general role respectively. \\
A DL-Lite$_{\textit{core}}$ ontology $\mathcal{O}=\langle\mathcal{T}, \mathcal{A}\rangle$ consists of a TBox $\mathcal{T}$ and an ABox $\mathcal{A}$.  $\mathcal{T}$ is formed by a finite set of concept inclusion assertions of the form $B \sqsubseteq \mathrm{C}$. $\mathcal{A}$ is formed by a finite set of membership assertions on atomic concepts and on atomic roles, of the form $A(a)$ and $P(a,b)$. DL-Lite$_{\mathcal{R}}$ extends DL-Lite$_{\text {core}}$ with role inclusion assertions of the form $R  \sqsubseteq E$ and DL-Lite$_{\mathcal{F}}$ extends DL-Lite$_{\textit{core}}$ with functionality on roles or on their inverses of the form (Funct $R$). \\ 
Here are some examples of common syntactic errors: \\
$^{-}TeachesTo$ is incorrect, for the misplaced inverse operator;
$\exists^{-}$ is incorrect, for the inverse operator applied to a quantifier;
$\exists Professor$ is incorrect, for the quantifier with a concept following;
$\exists$ is incorrect, for the quantifier missing a role following;
$\exists \exists TeachesTo$ is incorrect, for the redundant multiple quantifiers;
$TeachesTo\exists$ is incorrect, for the misplaced quantifiers;
$Professor\neg$ is incorrect, for the misplaced negation operator;
$Professor\sqcap$ is incorrect, for conjoining incomplete concepts;
$Professor\sqcap TeachesTo$ is incorrect, for conjoining a concept with a role;
$TeachesTo\sqcap HasTutor$ is incorrect, for conjoining roles directly;
$Professor \exists TeachesTo$ is incorrect, for missing conjunction operator.\\ \\
\setlength{\parindent}{0pt} \textcolor{blue}{Given DL-Lite Axioms: } \\
MaterialEntity $\sqsubseteq$ $\neg$PhysicalObject \\
$\exists$hasPerformer$\neg$ $\sqsubseteq$ MusicalExpression \\
Investigation $\sqsubseteq$ $\exists$hasPart \\ 
Protocol $\sqsubseteq$ $\neg$Investigation$\neg$ \\
\textcolor{gray}{(··· more context here ···)} \\ \\
\setlength{\parindent}{0pt} \textcolor{blue}{Answer: } 
\vspace{1mm}
\hrule

\vspace{1cm}

\setlength{\parindent}{0pt} \textbf{Sample answer for syntax checking:} \\  \vspace{-3mm}
\hrule 
\vspace{1mm}
\textcolor{gray}{(··· more context here ··· )} \\
Now, let’s analyze each given DL-Lite axiom for syntactic correctness: \\ \\
1. MaterialEntity $\sqsubseteq \neg$ PhysicalObject \\
This axiom states that MaterialEntity is a subclass of the negation of PhysicalObject.\\
\textcolor{green}{This is syntactically correct.} \\ \\
2. $\exists$hasPerformer$\neg$ $\sqsubseteq$ Performance $\sqcap$ MusicalExpression \\
Here, $\exists$hasPerformer$\neg$ seems to be intended as an existential quantification, but the syntax $\neg$ is misplaced.
Correct form should be something like $\exists hasPerformer$ $\sqsubseteq$ Performance $\sqcap$ MusicalExpression. \\
\textcolor{green}{Therefore, this is syntactically incorrect due to the misplacement of the negation.} \\ 
\textcolor{gray}{(··· more context here ··· )}
\vspace{1mm}
\hrule 

\vspace{1cm}

\setlength{\parindent}{0pt} \textbf{Prompt-\textit{NI} for subsumption of concepts or roles:} \\  \vspace{-3mm}
\hrule 
\vspace{1mm}
\textcolor{blue}{Task Description}: \\
There are a DL-Lite ontology and some logical implications, and your task is to determine whether each of these logical implications can be deduced from the given ontology. \\  \\
\setlength{\parindent}{0pt} \textcolor{blue}{Given Ontology :} \\
Ability $\sqsubseteq$ $\neg$Disability \\
Ability $\sqsubseteq$ $\neg$Device \\
Ability $\sqsubseteq$ $\exists$isAssistedBy \\
\textcolor{gray}{(··· more context here ···)} \\ \\
\setlength{\parindent}{0pt} \textcolor{blue}{Logical Implications:} \\
Achondroplasia $\sqsubseteq$ PhysicalDisability \\
Amputation $\sqsubseteq$ PhysicalDisability \\
AssistiveDevice $\sqsubseteq$ Device \\
Autism $\sqsubseteq$ MentalDisability \\
\textcolor{gray}{(··· more context here ···)} \\ \\
\setlength{\parindent}{0pt} \textcolor{blue}{Answer: } 
\vspace{1mm}
\hrule 

\vspace{1cm}

\setlength{\parindent}{0pt} \textbf{Prompt-\textit{WI} for subsumption of concepts or roles:} \\  \vspace{-3mm}
\hrule 
\vspace{1mm}
\textcolor{blue}{Task Description}: \\
There are a DL-Lite ontology and some logical implications, and your task is to determine whether each of these logical implications can be deduced from the given ontology.
 \\ 
 Here, you are provided with some reasoning rules:\\
 $\alpha=C_1 \sqsubseteq C_2,   \beta=C_2 \sqsubseteq C_3   \rightarrow  \beta_{\text{new}}=C_1 \sqsubseteq C_3$\\
$\alpha=R_1 \sqsubseteq R_2 ,  \beta=R_2 \sqsubseteq R_3   \rightarrow  \beta_{\text{new}}=R_1 \sqsubseteq R_3$\\ 
$ \alpha=C_1 \sqsubseteq C_2 ,  \beta=C_2 \sqsubseteq \neg C_3  \rightarrow \beta_{\text {new}}=C_1 \sqsubseteq \neg C_3 $\\
$\alpha=C_1 \sqsubseteq C_2,  \beta=C_3 \sqsubseteq \neg C_2 \rightarrow \beta_{new}=C_1 \sqsubseteq \neg C_3 $\\
$ \alpha=R_1 \sqsubseteq R_2, \beta=\exists R_2 \sqsubseteq \neg C \rightarrow \beta_{\text {new}}=\exists R_1 \sqsubseteq \neg C $\\
$\alpha=R_1 \sqsubseteq R_2, \beta=C \sqsubseteq \neg \exists R_2 \rightarrow \beta_{\text {new}}=\exists R_1 \sqsubseteq \neg C $\\
$ \alpha=R_1 \sqsubseteq R_2, \beta=C \sqsubseteq \neg \exists R_2^{-} \rightarrow \beta_{\text {new}}=\exists R_1^{-} \sqsubseteq \neg C $\\
$ \alpha=R_1 \sqsubseteq R_2, \beta=\exists R_2^{-} \sqsubseteq \neg C \rightarrow \beta_{\text {new}}=\exists R_1^{-} \sqsubseteq \neg C $\\
$ \alpha=R_1 \sqsubseteq R_2, \beta=R_2 \sqsubseteq \neg R_3  \rightarrow \beta_{\text {new}}=R_1 \sqsubseteq \neg R_3 $ \\
$ \alpha=R_1 \sqsubseteq R_2, \beta=R_3 \sqsubseteq \neg R_2  \rightarrow \beta_{\text {new }}=R_1 \sqsubseteq \neg R_3 $\\
$ one \ of \  the \ assertions  \ R \sqsubseteq \neg R,  \exists R \sqsubseteq \neg \exists R,  \exists R^{-} \sqsubseteq \neg \exists R^{-} 
 \rightarrow the \ other \ two$ \\ \\
\setlength{\parindent}{0pt} \textcolor{blue}{Given Ontology :} \\
Ability $\sqsubseteq$ $\neg$Disability \\
Ability $\sqsubseteq$ $\neg$Device \\
Ability $\sqsubseteq$ $\exists$isAssistedBy \\
\textcolor{gray}{(··· more context here ···)} \\ \\
\setlength{\parindent}{0pt} \textcolor{blue}{Logical Implications:} \\
Achondroplasia $\sqsubseteq$ PhysicalDisability \\
Amputation $\sqsubseteq$ PhysicalDisability \\
AssistiveDevice $\sqsubseteq$ Device \\
Autism $\sqsubseteq$ MentalDisability \\
\textcolor{gray}{(··· more context here ···)} \\ \\
\setlength{\parindent}{0pt} \textcolor{blue}{Answer: } 
\vspace{1mm}
\hrule 

\vspace{1cm}

\setlength{\parindent}{0pt} \textbf{Prompt-\textit{WIE} for subsumption of concepts or roles:} \\  \vspace{-3mm}
\hrule 
\vspace{1mm}
\textcolor{blue}{Task Description}: \\
There are a DL-Lite ontology and some logical implications, and your task is to determine whether each of these logical implications can be deduced from the given ontology.
 \\ 
 Here, you are provided with some reasoning rules:\\
 $\alpha=C_1 \sqsubseteq C_2,   \beta=C_2 \sqsubseteq C_3   \rightarrow  \beta_{\text{new}}=C_1 \sqsubseteq C_3$\\
$\alpha=R_1 \sqsubseteq R_2 ,  \beta=R_2 \sqsubseteq R_3   \rightarrow  \beta_{\text{new}}=R_1 \sqsubseteq R_3$\\ 
$ \alpha=C_1 \sqsubseteq C_2 ,  \beta=C_2 \sqsubseteq \neg C_3  \rightarrow \beta_{\text {new}}=C_1 \sqsubseteq \neg C_3 $\\
$\alpha=C_1 \sqsubseteq C_2,  \beta=C_3 \sqsubseteq \neg C_2 \rightarrow \beta_{new}=C_1 \sqsubseteq \neg C_3 $\\
$ \alpha=R_1 \sqsubseteq R_2, \beta=\exists R_2 \sqsubseteq \neg C \rightarrow \beta_{\text {new}}=\exists R_1 \sqsubseteq \neg C $\\
$\alpha=R_1 \sqsubseteq R_2, \beta=C \sqsubseteq \neg \exists R_2 \rightarrow \beta_{\text {new}}=\exists R_1 \sqsubseteq \neg C $\\
\textcolor{gray}{(··· more context here ···)} \\
Here are some examples: \\
If HasParent $\sqsubseteq$ HasAncestor and Mortal $\sqsubseteq$ $\neg\exists$HasAncestor, then $\exists$HasParent $\sqsubseteq$ $\neg$Mortal. \\
If HasParent $\sqsubseteq$ HasAncestor and Immortal $\sqsubseteq$ $\neg\exists$hasAncestor$^{-}$, then $\exists$hasParent$^{-}$ $\sqsubseteq$ $\neg$Immortal. \\
If HasParent $\sqsubseteq$ HasAncestor and $\exists$ hasAncestor$^{-}$ $\sqsubseteq$ $\neg$immortal, then $\exists$hasParent$^-$ $\sqsubseteq$ $\neg$Immortal.
If HasParent $\sqsubseteq$ HasAncestor and HasAncestor $\sqsubseteq$ $\neg$HasSibling, then HasParent $\sqsubseteq$ $\neg$HasSibling.\\
\textcolor{gray}{(··· more context here ···)} \\ \\
 
\setlength{\parindent}{0pt} \textcolor{blue}{Given Ontology :} \\
Ability $\sqsubseteq$ $\neg$Disability \\
Ability $\sqsubseteq$ $\neg$Device \\
Ability $\sqsubseteq$ $\exists$isAssistedBy \\
\textcolor{gray}{(··· more context here ···)} \\ \\

\setlength{\parindent}{0pt} \textcolor{blue}{Logical Implications:} \\
Achondroplasia $\sqsubseteq$ PhysicalDisability \\
Amputation $\sqsubseteq$ PhysicalDisability \\
AssistiveDevice $\sqsubseteq$ Device \\
Autism $\sqsubseteq$ MentalDisability \\
\textcolor{gray}{(··· more context here ···)} \\ \\
\setlength{\parindent}{0pt} \textcolor{blue}{Answer: } 
\vspace{1mm}
\hrule 

\vspace{1cm}

\setlength{\parindent}{0pt} \textbf{Sample answer for subsumption of concepts or roles:} \\  \vspace{-3mm}
\hrule 
\vspace{1mm}
\textcolor{gray}{(··· more context here ··· )} \\
1. Professor $\sqsubseteq$ Staff \\
Professor is a subclass of Academic\_Staff and Academic\_Staff is  a subclass of Staff in the ontology. \textcolor{green}{So, this implication holds}. \\ \\

2. BatteryPowered\_Wheelchair $\sqsubseteq$ Motorised\_Wheelchair\\
There's no direct assertion about BatteryPowered\_Wheelchair being a subclass of Motorised\_Wheelchair in the TBox. \textcolor{red}{Therefore, this implication cannot be deduced.}  \\

\textcolor{gray}{(··· more context here ··· )}
\vspace{1mm}
\hrule 

\vspace{1cm}

\setlength{\parindent}{0pt} \textbf{Prompt-\textit{NI} for instance checking:} \\  \vspace{-3mm}
\hrule 
\vspace{1mm}
\textcolor{blue}{Task Description}: \\
There are a DL-Lite ontology and some logical implications, and your task is to determine whether each of these logical implications can be deduced from the given ontology.
 \\  \\
\setlength{\parindent}{0pt} \textcolor{blue}{Given Ontology :} \\
\textcolor{gray}{(··· more context here ···)} \\
AssistantProfessor(AssistantProfessor0) \\
SportsFan(AssistantProfessor0) \\
\textcolor{gray}{(··· more context here ···)} \\ \\
\setlength{\parindent}{0pt} \textcolor{blue}{Logical Implications:} \\
Man(AssistantProfessor0) \\
SportsLover(AssistantProfessor0) \\
\textcolor{gray}{(··· more context here ···)} \\ \\
\setlength{\parindent}{0pt} \textcolor{blue}{Answer: } 
\vspace{1mm}
\hrule 

\vspace{1cm}

\setlength{\parindent}{0pt} \textbf{Prompt-\textit{WI} for instance checking:} \\  \vspace{-3mm}
\hrule 
\vspace{1mm}
\textcolor{blue}{Task Description}: \\
There are a DL-Lite ontology and some logical implications, and your task is to determine whether each of these logical implications can be deduced from the given ontology.
 \\ 
 Here, you are provided with some reasoning rules:\\
 $ \alpha=C_{1} \sqsubseteq C_{2}, \beta=C_{1}(a)  \rightarrow  \beta_{\text {new }}=C_{1}(a) $\\
$\alpha=C \sqsubseteq \exists R , \beta=C(a)  \rightarrow  \beta_{\text {new }}=R\left(a, a_{\text {new}}\right) $\\
$ \alpha=\exists R \sqsubseteq C, \beta=R\left(a, a^{\prime}\right)  \rightarrow  \beta_{\text {new }}=C(a) $\\
$\alpha=\exists R_{1} \sqsubseteq \exists R_{2}, \beta=R_{1}\left(a, a^{\prime}\right)  \rightarrow  \beta_{\text {new }}=R_{2}\left(a, a_{\text {new }}\right)$ \\
$ \alpha=R_{1} \sqsubseteq R_{2},  \beta=R_{1}\left(a, a^{\prime}\right)  \rightarrow  \beta_{\text {new }}=R_{2}\left(a, a^{\prime}\right)$  \\ \\
\setlength{\parindent}{0pt} \textcolor{blue}{Given Ontology :} \\
AssistantProfessor(AssistantProfessor0) \\
SportsFan(AssistantProfessor0) \\
\textcolor{gray}{(··· more context here ···)} \\ \\
\setlength{\parindent}{0pt} \textcolor{blue}{Logical Implications:} \\
Man(AssistantProfessor0) \\
SportsLover(AssistantProfessor0) \\
\textcolor{gray}{(··· more context here ···)} \\ \\
\setlength{\parindent}{0pt} \textcolor{blue}{Answer: } 
\vspace{1mm}
\hrule 

\vspace{1cm}

\setlength{\parindent}{0pt} \textbf{Prompt-\textit{WIE} for instance checking:} \\  \vspace{-3mm}
\hrule 
\vspace{1mm}
\textcolor{blue}{Task Description}: \\
There are a DL-Lite ontology and some logical implications, and your task is to determine whether each of these logical implications can be deduced from the given ontology.
 \\ 
 Here, you are provided with some reasoning rules:\\
 $ \alpha=C_{1} \sqsubseteq C_{2}, \beta=C_{1}(a)  \rightarrow  \beta_{\text {new }}=C_{1}(a) $\\
\textcolor{gray}{(··· more context here ···)}   \\ 
Here are examples: \\
If Human $\sqsubseteq$ Animal and Human(John), then Animal(John) \\
If Human $\sqsubseteq$ $\exists$hasParent and Human(John), then hasParent(John, \_). \\
If $\exists$hasChild $\sqsubseteq$ Parent and HasChild(John, \_), then Parent(John). \\
If $\exists$hasChild $\sqsubseteq$ $\exists$hasDescendant and HasChild(John, \_), then hasDescendant(John, \_). \\
If HasParent $\sqsubseteq$ HasRelative and HasParent(Mary, John), then HasRelative(Mary, John). \\ \\
\setlength{\parindent}{0pt} \textcolor{blue}{Given Ontology :} \\
AssistantProfessor(AssistantProfessor0) \\
SportsFan(AssistantProfessor0) \\
\textcolor{gray}{(··· more context here ···)} \\ \\
\setlength{\parindent}{0pt} \textcolor{blue}{Logical Implications:} \\
Man(AssistantProfessor0) \\
SportsLover(AssistantProfessor0) \\
\textcolor{gray}{(··· more context here ···)} \\ \\
\setlength{\parindent}{0pt} \textcolor{blue}{Answer: } \
\vspace{1mm}
\hrule 

\vspace{1cm}

\setlength{\parindent}{0pt} \textbf{Sample answer for instance checking:} \\  \vspace{-3mm}
\hrule 
\vspace{1mm}
\textcolor{gray}{(··· more context here ··· )} \\
1. Man(AssistantProfessor0) - \textcolor{green}{True}, since AssistantProfessor0 is explicitly stated to be a Man.
 \\ 
\textcolor{gray}{(··· more context here ··· )}
\vspace{1mm}
\hrule 

\vspace{1cm}

\setlength{\parindent}{0pt} \textbf{Prompt for property characteristic probing:} \\  \vspace{-3mm}
\hrule 
\vspace{1mm}
\textcolor{blue}{Task Description}: \\
There are a DL-Lite ontology and some logical implications, and your task is to determine whether each of these logical implications can be deduced from the given ontology. \textcolor{purple}{Give reasons or inferring process.}
 \\  \\
\setlength{\parindent}{0pt} \textcolor{blue}{Given Ontology :} \\
\textcolor{gray}{(··· more context here ···)} \\
WorksIn$^{-}$ $\sqsubseteq$ Employs \\
Employs $\sqsubseteq$ WorksIn$^{-}$ \\
(funct WorksIn) \\
WorksIn(John, Google)  \\
WorksIn(John, x) \\
\textcolor{gray}{(··· more context here ···)}  \\ \\ 
\setlength{\parindent}{0pt} \textcolor{blue}{Logical Implications:} \\
Employs(Google, John) \\
x $\equiv$ John \\
\textcolor{gray}{(··· more context here ···)} \\ \\
\setlength{\parindent}{0pt} \textcolor{blue}{Answer: }
\vspace{1mm}
\hrule 

\vspace{1cm}

\setlength{\parindent}{0pt} \textbf{Sample answer for property characteristic probing:} \\  \vspace{-3mm}
\hrule 
\vspace{1mm}
\textcolor{gray}{(··· more context here ··· )} \\ \\
1. Employs(Google, John) - \textcolor{green}{True}, Since  WorksIn(John, Google) is given  and  Employs is the inverse of WorksIn, hence Employs(Google, John) can be deduced.
 \\ \\
2. x $\equiv$ John - \textcolor{green}{True},
 Given: 
WorksIn(John, Google)  and 
WorksIn(John, x). Since WorksIn is a functional property, John can only work at one company. Thus, x must be Google to satisfy the functional constraint.\\ \\
\textcolor{gray}{(··· more context here ··· )}
\vspace{1mm}
\hrule 

\vspace{1cm}

\setlength{\parindent}{0pt} \textbf{Prompt for query answering:} \\  \vspace{-3mm}
\hrule 
\vspace{1mm}
\textcolor{blue}{Task Description}: \\
There are a DL-Lite ontology and a query, and your task is to answer the query over the given DL-Lite ontology. \textcolor{purple}{Because ontology is relatively large, it will be entered in several times.}
\\ \\
\setlength{\parindent}{0pt} \textcolor{blue}{Given Ontology :} \\
\textcolor{gray}{(··· more context here ···)} \\
PhDStudent $\sqsubseteq$ Student \\
Student $\sqsubseteq$ $\neg$$\exists$hasStaffID \\
$\exists$hasStaffID$^{-}$ $\sqsubseteq$ ID \\
Student $\sqsubseteq$ $\neg$ID \\
PhdStudent(John) \\
\textcolor{gray}{(··· more context here ···)} \\ \\
\setlength{\parindent}{0pt} \textcolor{blue}{Queries:} \\
Q1(x) $\leftarrow$ Student(x) \\
\textcolor{gray}{(··· more context here ···)} \\ \\
\setlength{\parindent}{0pt} \textcolor{blue}{Answer: }
\vspace{1mm}
\hrule 

\vspace{1cm}

\setlength{\parindent}{0pt} \textbf{Sample answer for query answering:} \\  \vspace{-3mm}
\hrule 
\vspace{1mm}
\textcolor{gray}{(··· more context here ··· )} \\ \\
1. From axiom 5, we know PhDStudent(John). 
From axiom 1, we have PhDStudent $\sqsubseteq$
Student. This means every PhDStudent is a Student. Therefore, PhDStudent(John) implies 
Student(John). The answer is \textcolor{green}{q(John)}.
 \\ \\
\textcolor{gray}{(··· more context here ··· )}
\vspace{1mm}
\hrule 

\vspace{1cm}

\setlength{\parindent}{0pt} \textbf{Prompt for ontology satisfiability checking:} \\  \vspace{-3mm}
\hrule 
\vspace{1mm}
\textcolor{blue}{Task Description}: \\
There is a DL-Lite ontology, and your task is to determine whether the ontology is satisfiable, i.e., iff it has at least one model.
\\ \\
\setlength{\parindent}{0pt} \textcolor{blue}{Given Ontology :} \\
\textcolor{gray}{(··· more context here ···)} \\
MasterStudent $\sqsubseteq$ Student \\
MasterStudent $\sqsubseteq$ Employee \\
Student $\sqsubseteq$ $\neg$ Employee \\
MasterStudent(John) \\
\textcolor{gray}{(··· more context here ···)} \\ \\
\setlength{\parindent}{0pt} \textcolor{blue}{Answer: }
\vspace{1mm}
\hrule 

\vspace{1cm}

\setlength{\parindent}{0pt} \textbf{Sample answer for ontology satisfiability checking:} \\  \vspace{-3mm}
\hrule 
\vspace{1mm}
\textcolor{gray}{(··· more context here ··· )} \\ \\
The axioms lead to a logical inconsistency regarding the concept of 
MasterStudent
MasterStudent. Therefore, the given DL-Lite ontology is \textcolor{green}{not satisfiable}.
 \\ \\
\textcolor{gray}{(··· more context here ··· )}
\vspace{1mm}
\hrule 

\vspace{1cm}

\section{Instructions about Protégé} \label{AD}
We import the selected ontological datasets into Protégé and utilize the reasoning engine 
HermiT 1.3.8.413 to infer logical implications. 
\begin{figure}[htbp]
    \centering
    \includegraphics[scale=0.56]{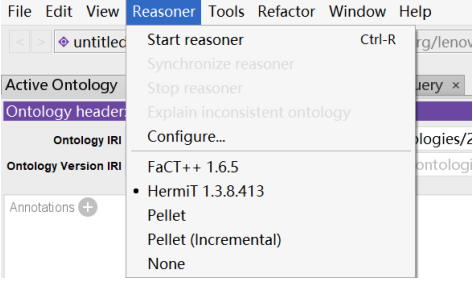}
\end{figure} 

Then we export the inferred axioms. For  subsumption of concepts or roles, the chosen 
categories of inferred axioms exported are subclasses, sub object properties, and sub data properties. For instance checking, the chosen categories of inferred axioms exported are class assertions and property assertions
\begin{figure}[htbp]
    \centering
    \includegraphics[scale=0.46]{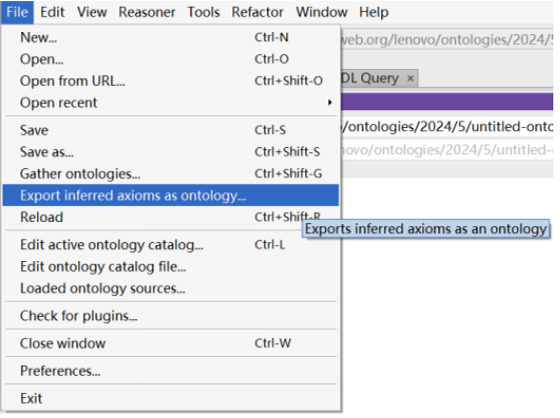}
\end{figure} 
\begin{figure}[htbp]
    \centering
    \includegraphics[scale=0.42]{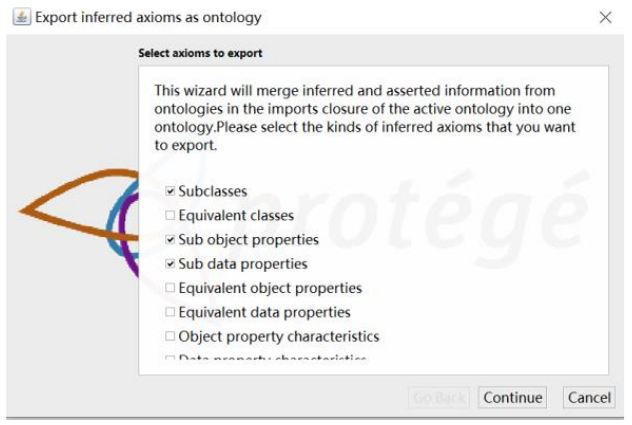}
\end{figure}

\end{document}